\documentclass[10pt,journal,cspaper,compsoc,twocolumn,twoside]{IEEEtran}
\usepackage{amssymb,amsmath}
\usepackage{balance}
\usepackage{epsfig}
\usepackage{algorithm} 
\usepackage{algorithmic} 
\usepackage{booktabs}
\usepackage{graphicx,subfigure}
\usepackage{graphics}
\usepackage{array}
\usepackage{multirow}
\usepackage{flushend}
\usepackage{color}
\usepackage{amsthm}
\newtheorem{myDef}{Definition}

\makeatletter

\newcommand{\Rmnum}[1]{\expandafter\@slowromancap\romannumeral #1@}
\makeatother
\ifCLASSOPTIONcaptionsoff
  \usepackage[nomarkers]{endfloa t}
 \let\MYoriglatexcaption\caption
 \renewcommand{\caption}[2][\relax]{\MYoriglatexcaption[#2]{#2}}
\fi

\begin{document}
%
\title{Personalized Age Progression with Bi-level Aging Dictionary Learning}
%
%

\author{Xiangbo~Shu, Jinhui~Tang, \textit{Senior Member, IEEE}, Zechao~Li, Hanjiang~Lai, Liyan~Zhang \\ and Shuicheng~Yan, \textit{Fellow, IEEE}
\thanks{X. Shu, J. Tang and Z. Li are with the School of Computer Science and Engineering, Nanjing
University of Science and Technology, Nanjing, 210094, China (e-mail: shuxb@njust.edu.cn, jinhuitang@njust.edu.cn and zechao.li@njust.edu.cn). J. Tang is the corresponding author.}
\thanks{H. Lai is with School of data and computer science, Sun Yat-sen university, Guangzhou, 510275, China (e-mail: laihanj@gmail.com).}
\thanks{L. Zhang is with is with the College of Computer Science and Technology, Nanjing University of Aeronautics and Astronautics, Nanjing, 210016, China, and also with the Collaborative Innovation Center of Novel Software Technology and Industrialization, Nanjing, 210023, China (e-mail: zhangliyan@nuaa.edu.cn).}
\thanks{S. Yan is with the Department of Electrical and Computer Engineering, National University of Singapore, 117576, Singapore (e-mail: eleyans@nus.edu.sg).}}

%
%

\markboth{IEEE~TRANSACTIONS~ON~PATTERN~ANALYSIS~AND~MACHINE~INTELLIGENCE, 2017}%
{IEEE~TRANSACTIONS~ON~PATTERN~ANALYSIS~AND~MACHINE~INTELLIGENCE, 2017}
%


\IEEEcompsoctitleabstractindextext{%
\begin{abstract}

Age progression is defined as aesthetically re-rendering the aging face at any future age for an individual face. In this work, we aim to automatically render aging faces in a personalized way. Basically, for each age group, we learn an aging dictionary to reveal its aging characteristics (e.g., wrinkles), where the dictionary bases corresponding to the same index yet from two neighboring aging  dictionaries form a particular aging pattern cross these two age groups, and a linear combination of all these patterns expresses a particular personalized aging process. Moreover, two factors are taken into consideration in the dictionary learning process. First, beyond the aging dictionaries, each person may have extra personalized facial characteristics, e.g. mole, which are invariant in the aging process. Second, it is challenging or even impossible to collect faces of all age groups for a particular person, yet much easier and more practical to get face pairs from neighboring age groups. To this end, we propose a novel Bi-level Dictionary Learning based Personalized Age Progression (BDL-PAP) method. Here, bi-level dictionary learning is formulated to learn the aging dictionaries based on face pairs from neighboring age groups. Extensive experiments well demonstrate the advantages of the proposed BDL-PAP over other state-of-the-arts in term of personalized age progression, as well as the performance gain for cross-age face verification by synthesizing aging faces.
\end{abstract}

}

\maketitle

\IEEEdisplaynotcompsoctitleabstractindextext

%
\IEEEpeerreviewmaketitle

\section{Introduction}
\label{Introduction}
Age progression~\cite{ramanathan2009age}, also called age synthesis~\cite{fu2010age} or
face aging~\cite{suo2010compositional}, is defined as aesthetically rendering a face image with natural aging and rejuvenating effects for an individual face. It has been widely applied to various application domains, e.g., cross-age face analysis~\cite{park2010age}, authentication systems, finding lost children, entertainment, etc. There are two main categories of solutions to the age progression task: prototyping-based age progression~\cite{kemelmacher2014illumination,tiddeman2001prototyping,gandhi2004method} and modeling-based age progression~\cite{suo2010compositional,tazoe2012facial,liang2011multi}. Prototyping-based age progression transfers the difference between two prototypes (e.g., average faces) of the pre-divided source age group and the target age group into the input individual face, of which its age belongs to the source age group. Modeling-based age progression models the facial parameters of different ages (age ranges) for the shape and texture synthesis.

Intuitively, the natural aging process of a specific person usually follows the general rules of the human aging process. Meanwhile, the natural aging face of a specific person also contains some personalized facial characteristics, e.g., mole, birthmark, etc., which are almost invariant with time. Generally, prototyping-based age progression methods cannot well preserve this personality of an individual face, since they are based on the general rules of the human aging process for a relatively large population. And modeling-based age progression methods do not specially consider the personalized details for a specific person. Moreover, they require a large number of dense long-term (e.g., age span exceeds 20 years) face aging sequences for building the complex functions. However, collecting these required sequences in the real world is very difficult or even impossible. Fortunately, we have observed that the short-term (e.g., age span smaller than 10 years) face aging sequences are available in the Internet, such as photos of celebrities at different ages on Facebook/Twitter. Some available face aging datasets~\cite{chen2014cross,fgnet,ricanek2006morph} also contain the dense short-term sequences. Therefore, generating personalized age progression for an individual face by leveraging short-term face aging sequences is more feasible.

\begin{figure*}[t]
	\centering
	\includegraphics[scale=0.62]{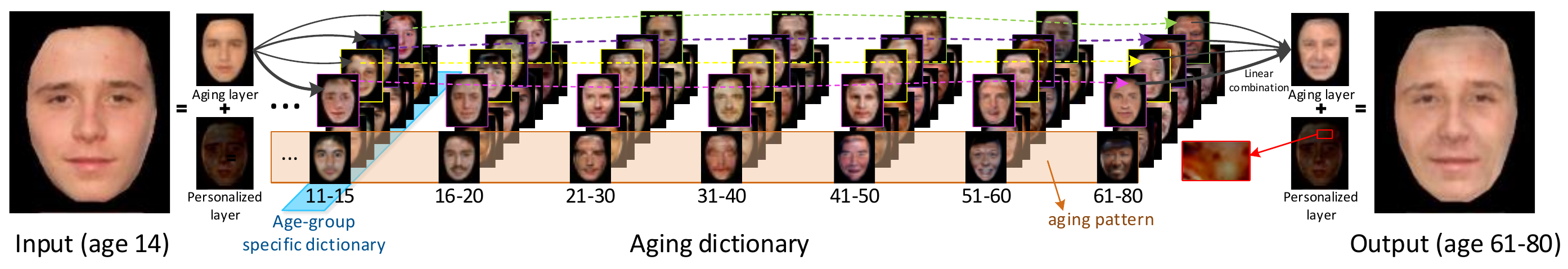}
	\caption{A personalized aging face generated by the proposed method. This aging face contains the aging layer (e.g., wrinkles) and the personalized layer (e.g., mole). The former can be seen as the corresponding face in a linear combination of the aging patterns, while the latter is invariant in the aging process. Better view in color.}
	\label{fig0}
\end{figure*}

In this work, we propose a Bi-level Dictionary Learning based Personalized Age Progression (BDL-PAP) method, which leverages short-term face aging pairs to automatically render aging faces in a personalized way, as shown in Figure~\ref{fig0}.
Primarily, based on the aging-(in)variant patterns in the face aging process, an individual face can be decomposed into an aging layer and a personalized layer. The former shows the general aging characteristics (e.g., wrinkles), while the latter shows the personalized facial characteristics (e.g., mole). For different human age groups (e.g., 11-15, 16-20, etc), we design the corresponding aging dictionaries to reveal the general aging characteristics, where the dictionary bases with the same index yet from two neighboring aging dictionaries form a particular aging patterns (e.g., they are linked by a dotted line in Figure \ref{fig0}). Therefore, the aging layer of the aging face can be reconstructed by a linear combination of several aging dictionary bases with sparse representation. The motivation of the sparsity assumption is to use fewer dictionary bases for reconstruction such that the reconstructed aging layer of face can be shaper and less blurred~\cite{wright2009robust,shu2016image}. The residual between the reconstructed  aging layer and the input face can be defined as the personalized layer, which is invariant in the aging process. Finally, we render the aging face in a future age group for an input face by fusing the reconstructed aging layer of this age group and the personalized layer.

\begin{figure*}[!t]
	\centering
	\includegraphics[scale=0.45]{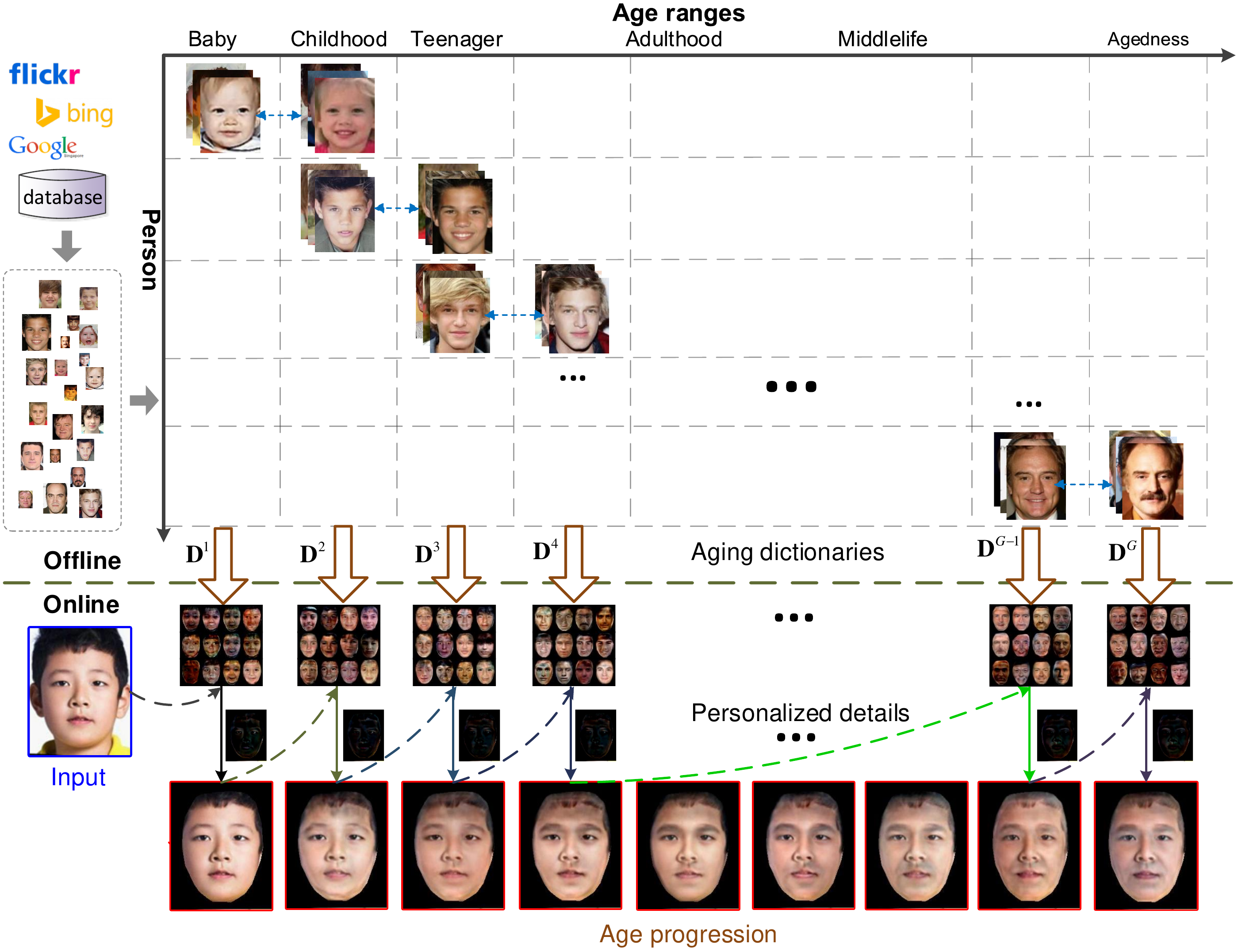}
	\caption{Framework of the proposed age progression. ${\bf D}^{g}$ denotes a aging dictionary of the $g$-th age group. In the offline phase, we collect short-term aging face pairs and then train the aging dictionary. In the online phase, for an input face, we render its aging faces by bi-level optimization on the corresponding aging dictionaries. }
	\label{fig1}
\end{figure*}

To learn the aging dictionaries, we propose a bi-level dictionary learning model, and utilize the more practical short-term face aging pairs as the training set instead of the possibly unavailable long-term face aging sequences. In the bi-level dictionary learning, we assume that the sparse representation of a younger-aging layer of one face w.r.t. the younger-aging dictionary can directly reconstruct its older-aging layer using the older-aging dictionary. The distribution of the collected face aging pairs in this work is shown in the upper part of Figure~\ref{fig1}. We can see that: 1)~each age group has its own aging dictionary; 2)~every two neighboring age groups are linked by several dense short-term face aging pairs,  which makes all the age
groups  linked  together; 3)~the personalized layer contains the personalized facial characteristics. These three properties are able to guarantee that all aging dictionaries can be trained well by the bi-level dictionary learning on the short-term face aging pairs.

The main contributions of this work are summarized as
follows:
\begin{itemize}
\item We propose a Personalized Age Progression method based on the Bi-level Dictionary Learning (BDL-PAP) to render aging faces, which can preserve the personalized facial characteristics.
\item  Since it is challenging or even impossible to collect intra-person face sequences of all age groups, the proposed method only requires the easy-acquired short-term face aging pairs to learn all aging dictionary bases of human aging, which is more feasible.
\item Extensive experiments well validate the advantage of the proposed solution over other state-of-the-arts in term of personalized aging progression, as well as the performance gain for cross-age face verification by synthesizing aging faces.
\end{itemize}

Compared to our preliminary work in~\cite{shu2015personalized}, we have the following extensions in this paper: 1)~we extend the preliminary coupled dictionary learning model to a novel bi-level dictionary learning model, which leads to a more effective and efficient personalized age progression method; 2)~we compare the proposed age progression method with more state-of-the-art methods by conducting age progression experiments and cross-age face verification experiments; 3)~we illustrate the effectiveness of the proposed personalized age progression method by comparing to its unpersonalized version; 4)~we show the efficiency of the proposed method by comparing the executing time of different age progression methods.

The rest of this paper is organized as follows. Section~\ref{RW} reviews related works. Section~\ref{PAPBAD} details the proposed method, including the whole framework, dictionary learning model, objective function, optimization procedure, and age progression synthesis. Experiments are conducted in Section~\ref{Exe}. Finally, Section~\ref{C} concludes this work and discusses the future works.

\section{Related Work}
\label{RW}

\subsection{Age Progression  Methods}

Age progression has been comprehensively reviewed in literature~\cite{fu2010age,ramanathan2009age,ramanathan2009computational,shu2016age}.
As one of the early studies, Burt et al.~\cite{burt1995perception} focused on creating average faces for different ages and transferring the facial difference between the average faces into the input face. This method gives an insight into the age progression task. Thereafter, some prototyping-based aging methods~\cite{kemelmacher2014illumination,tiddeman2001prototyping,shu2016kinship} are proposed to improve the age progression method in~\cite{burt1995perception}. Kemelmacher et al.~\cite{kemelmacher2014illumination} proposed an Illumination-Aware Age Progression (IAAP) method by leveraging the difference between the warped average faces instead of the original average faces. Here the warped average face is computed based on the flow from the average face to the input face. Generally, the aging progression of an individual is stochastic and
non-deterministic in the time dimension. Therefore, Shu et al.~\cite{shu2016kinship} presented a kinship-guided age progression approach to automatically generate aging faces by leveraging kinship information.
The  aging faces generated by these methods almost have no personalized characteristic and the aging speed of different people is synchronous. Although some researchers tried to consider the individual-specific face age progression~\cite{solomon2006person,hubball2008image,hunter2009synthesis}, lack of personality is still a challenging problem.

Modeling-based age progression is the other type of the age progression, which considers shape and texture synthesis simultaneously~\cite{albert2007review}. Some modeling-based age progression methods have been proposed, including active appearance model~\cite{lanitis2002toward}, craniofacial growth model~\cite{ramanathan2006face}, and-or graph model~\cite{suo2010compositional}, statistical model~\cite{paysan2010statistical}, implicit function~\cite{berg2006facial,schroeder2007facial}, etc. Generally, to model large appearance changes over a long-term face aging sequence, modeling-based age progression requires sufficient training data. Suo et al.~\cite{suo2012concatenational} attempted to learn long-term aging patterns from available short-term aging databases by a proposed concatenational graph evolution aging model. Recent years, inspired by the success of Recurrent Neural Network
 (RNN), Wang et al.~\cite{wang2016recurrent} exploited RNNs to model the whole aging sequence. The bottom layer in a RNN works as an encoder, which projects the image to a high-dimension space. The top layer in a RNN works as a decoder, which decodes the hidden representation to an aging face.


\subsection{Age Progression based Face Analysis }
Face analysis (e.g., face verification and face recognition)~\cite{sun2014deep,schroff2015facenet,ding2016multi} has achieved great progress in the last decade. Usually, as a natural biological process, face aging can change the appearance and texture of the facial landmarks, such as wrinkles, senile plaque, mustache, etc. Therefore, the performance of existing face analysis methods for the cross-age face analysis will degree since the age gap exists between the two faces of one person at different ages. By conducting experiments on a passport photo database, Ling et al.~\cite{ling2007study} validated that the face aging would increase the challenge of face recognition. Towards this end,  some methods are proposed to address the cross-age face analysis problem in recent years~\cite{chen2014cross,guo2010cross,wang2006age,park2010age,park2008face,ling2007study,gong2015maximum,wu2012age,lanitis2009survey,ling2010face}. These methods can be divided into two categories. Methods of one category aim to learn the age-invariant features for cross-age face analysis~\cite{gong2015maximum,ling2007study,chen2014cross,gong2013hidden}. For example, Ling et al.~\cite{ling2007study} found that the gradient direction is insensitive to the face aging, and then proposed to use the gradient directions of pyramid structure as the feature descriptors of face images. Gong et al.~\cite{gong2013hidden} proposed to separate the face pattens into the age-invariant and age-variant patterns, and then use the age-invariant patterns as the inputs of learning model. We can conclude that how to learn the age-invariant features or extract the age-invariant patterns of face images is the key point. Methods of the other category are proposed to use the age synthesis  to eliminate the age gap~\cite{wu2012age,wang2006age,park2008face,park2010age,patterson2006automatic,suo2009learning,suo2010compositional,suo2012concatenational,liu2016deep}. Taking the face verification as an example, for two faces of one person at different ages, the age progression method renders a synthesized aging face from the younger face. And then, the synthesized face instead of the original face is used to implement the face verification.

\subsection{Face Aging Datasets}
FG-NET~\cite{fgnet} is one popular face aging dataset. It has been used to evaluate the age estimation~\cite{geng2007automatic,guo2009human}, cross-age face verification, cross-age face recognition~\cite{chang2011ordinal,park2008face}, and age progression~\cite{kemelmacher2014illumination,suo2012concatenational,suo2010compositional,suo2007multi}. This dataset contains 1,002 face photos from 82 persons within 0-69 age range: about 64\%
of the images are from children (with ages $<$ 18), and around 36\% are from adults (with ages $\geqslant$ 18). All photos are taken by digital camera and film-based camera. Morph~\cite{ricanek2006morph} is one of the largest face aging dataset. It has been also used to evaluate the age estimation, age progression, and face recognition. It is composed of Album1 subset and Album2 subset. The Album1 subset contains 1,690 face images from 515 persons. And the Album2 subset contains 94,000 face images from 24,000 people, which are collected from different places. In Morph, the size of all images is resized to $200\times240$ or $400\times480$, and each image has the corresponding age, gender and race labels. Recently, about 55,000 face images from 13,000 people in Morph are available on the website. In recent years, Chen et al.~\cite{chen2014cross} released a new Cross-Age Celebrity Dataset (CACD), which is collected from image search websites. CACD contains 163,446 face images from 2,000 celebrities within 16-62 age range. The age label of each face photo is estimated by referring to its shooting time, namely the shooting time is the age label. Although these age labels are inaccurate, the relative age for the same person is accurate. Thus, this dataset can be used for the cross-age face search and recognition. 

\section{The Proposed Method}
\label{PAPBAD}

\subsection{Overview of the Framework}
\label{OOF}
The framework of the proposed age progression method is shown in Figure~\ref{fig1}. The offline phase is described as follows. First, we collect the dense short-term aging pairs of the same persons from Internet and also from available datasets. Second, for each predefined age group, we learn a corresponding aging dictionary to represent its aging characteristics by the proposed bi-level dictionary learning model. In the online phase, for an input face, we generate its aging faces step by step, from the current age to the target age. Specifically, we first generate the aging face in the next age group by the corresponding aging dictionary with an implicitly common sparse representation, as well as a personalized layer. After that, taking this new aging face as
the input, and repeat
the above process until all aging faces have been rendered.

\subsection{Coupled Dictionary Learning}
We divide the human age range into $G$ age groups (each group spans less than 10 years) in this work. Let $\{{\bf x}_i^1,\cdots,{\bf x}_i^g,\cdots,{\bf x}_i^G\}$ denote a selected face aging sequence of the $i$-th person covering all age groups, where $i=1,2,\cdots,L$ ($L$ is the number for the persons). Here, the face photo ${\bf x}_i^g\in \mathbb{R}^f$ falls into the $g$-th age group, where $f$ is the number of pixels in a face photo. For the $g$-th age group ($g=1,2,\cdots,G$), we define its aging dictionary ${\bf B}^g$ to capture the aging characteristics, which will be learned in the following. 

{\bf Personality-aware formulation.} The aging dictionary learning in this work considers the personalized details of an individual when representing the face aging sequences on their own aging dictionaries. Since the personalized characteristics are aging-invariant, such as mole, birthmark, permanent scar, etc., we plan to add a personalized layer  ${\bf p}_i\in \mathbb{R}^{f}$ for a face aging sequence $\{{\bf x}_i^1,{\bf x}_i^2,\cdots,{\bf x}_i^G\}$ to indicate the personalized details of the $i$-th person in the aging process. Moreover, considering the computational efficiency, we utilize PCA to reduce the dimension of the dictionary. Let ${\bf H}^g\in \mathbb{R}^ {f \times m}$ denote a PCA projected matrix of all data in the $g$-th age group, and we have ${\bf B}^g={\bf H}^g{\bf D}^g$. Thus the original aging dictionary ${\bf B}^g$ is redefined as ${\bf D}^g\in \mathbb{R}^ {m \times k}$, where $k$ denotes the number of dictionary bases. All aging dictionaries compose an overall aging dictionary ${\bf D}=[{\bf D}^1,{\bf D}^2,\cdots,{\bf D}^G]\in \mathbb{R}^{m\times K}$, where $K=k\times  G$. So far, the aging face ${\bf \hat x}_i^{g+t}$ of ${\bf x}_i^g$
equals the linearly weighted combination of the aging dictionary bases in the ($g+t$)-th age group and the personalized layer ${\bf p}_i$, i.e., ${\bf \hat x}_i^{g+t}\approx{\bf H}^{g+t}{\bf D}^{g+t}{\bf a}_i+{\bf p}_i$ for $t=1,\cdots, G-g$, where ${\bf a}_i$ and ${\bf p}_i$ are the common sparse representation and the personalized layer, respectively. For $L$ face aging sequences $\{{\bf x}_i^1,\cdots,{\bf x}_i^G\}_{i=1}^{L}$ covering all age groups, a personality-aware dictionary learning model is formulated as follows,
\begin{equation} \label{eq1}
\begin{aligned}
&\mathop {\min }\limits_{\scriptstyle ~{\{\bf D}^{g}\}_{g=1}^G,\hfill \atop
	\scriptstyle {\{{\bf{a}}_i,{\bf{p}}_i\}_{i=1}^L} \hfill} \!\!\sum\limits_{g = 1}^{G} \!\sum\limits_{i = 1}^{L} \!\!\left\{ \!{\left\| {{{\bf{x}}_i^g} \!-\!\!{{\bf{H}}^g} {{\bf{D}}^g}{{\bf{a}}_i} \!\!-\!\! {{\bf{p}}_i}} \right\|_2^2 \!\!+ \!\!  \gamma\! \left\| {{{\bf{p}}_i}} \right\|_2^2\!+\! \lambda_1 {{\left\| {{{\bf{a}}_i}} \right\|}_1}  \!}\!\right\}\\
& ~~~~~\text{s.t.}{\kern 2pt}  {\left\| {{{\bf{D}}^g(:,d)}} \right\|_2} \le 1,\forall  d \!\in\! \{ 1, \cdots ,k\},\forall  g \!\in\! \{ 1, \cdots ,G\},
\end{aligned}
\end{equation}
where ${\bf D}^{g}(:,d)$ denotes the $d$-th column (base) of ${\bf D}^g$, $\lambda_1$ and $\gamma$ control the sparsity penalty and regularization, respectively. ${\bf D}^{g}(:,d)$ is used to represent the specific aging characteristics in the $g$-th age group.

{\bf Short-term coupled learning.} We observe that one person always has the dense short-term face aging photos, but no long-term face aging photos covering all age groups. Collecting these long-term face aging sequences in the real world is extremely difficult or even unlikely. Therefore, we have to use the shot-term face aging pairs instead of the long-term face sequences. Here, each face aging pair includes a certain person's two face images spanning two neighboring age groups. Let ${\mathcal{S}}^g$ ($g=1,2,\cdots,G$) denote a set of face images in the $g$-th age group, we assume that ${\mathcal{S}}^g$ and ${\mathcal{S}}^{g+1}$ share $n^g$ face pairs from $n^g$ different persons. Formally, we use $\{{\bf x}_{i^g} \!\in\! {\mathcal{S}}^g, {\bf y}_{i^{g}}\!\in\! {\mathcal{S}}^{g+1}\}$ to denote the face pair of the ${i^g}$-th person, where $i^g=1,2,\cdots, n^g$ is a local index related to the $g$-th age group. For the face aging pairs $\{{\bf x}_{i^g},{\bf y}_{i^g}\}_{i^g=1}^{n^g}$ spanning the $g$-th and ($g+1$)-th age groups, we reformulate a coupled dictionary learning model to simultaneously learn all aging dictionaries, i.e.,
\begin{equation} \label{eq2}
\begin{aligned}
\!\!\mathop {\min }\limits_{\scriptstyle ~~~~~~{\bf D}^{g},\hfill \atop
	\scriptstyle {\{{\bf{a}}_{i^g},{\bf{p}}_{i^g}\}_{i^g=1}^{n^g}} \hfill} \!\!&\sum\limits_{g = 1}^{G - 1} \sum\limits_{i^g = 1}^{n^g} \!\left\{ \!{\left\| {{{\bf{x}}_{i^g}} \!-\!{{\bf{H}}^g} {{\bf{D}}^g}{{\bf{a}}_{i^g}} \!-\! {{\bf{p}}_{i^g}}} \right\|_2^2 \!+ \!  \gamma\! \left\| {{{\bf{p}}_{i^g}}} \right\|_2^2  }\right.\\
\!\!&\left. +{\left\| {{{\bf{y}}_{i^g}} \!-\! {{\bf{H}}^{g + 1}}{{\bf{D}}^{g + 1}}{{\bf{a}}_{i^g}} \!-\! {{\bf{p}}_{i^g}}} \right\|_2^2  \!+\! \lambda_1 {{\left\| {{{\bf{a}}_{i^g}}} \right\|}_1}}\!\right\}  \\
\!\!\text{s.t.}{\kern 2pt}  || {\bf{D}}^g (:,d) &||_2 \!\le\! 1,\forall  d \!\in\! \{ 1, \cdots ,k\},\forall  g \!\in\! \{ 1, \cdots ,G\}.
\end{aligned}
\end{equation}

In Eq.~\eqref{eq2}, every two neighboring aging dictionaries ${\bf D}^g$ and ${\bf D}^{g+1}$ corresponding to two age groups are implicitly coupled via the common sparse representation ${{\bf{a}}_{i^g}}$, and the personalized layer ${{\bf{p}}_{i^g}}$ is to capture the personalized details of the  $i^g$-th person in the $g$-th age group, who has the face pair $\{{\bf x}_{i^g},{\bf y}_{i^g}\}$ spanning the $g$-th and ($g+1$)-th age groups. Let ${\bf D}\!=\![{\bf D}^1,\cdots,{\bf D}^G]\in \mathbb{R}^{m \times K}$, ${\bf X}^g\!=\![{\bf x}_{1^g},\cdots,{\bf x}_{n^g}]\in \mathbb{R}^{f\times n^g}$, ${\bf Y}^g\!=\![{\bf y}_{1^g},\cdots,{\bf y}_{n^g}]\in \mathbb{R}^{f\times n^g}$,  ${\bf P}^g=[{\bf p}_{1^g},\cdots,{\bf p}_{n^g}]\in \mathbb{R}^{f\times n^g}$, and ${\bf A}^g\!=\![{\bf a}_{1^g},\cdots,{\bf a}_{n^g}]\in \mathbb{R}^{k\times n^g}$, and Eq.~\eqref{eq2} can be rewritten in the matrix form
\begin{equation} \label{eq3}
\begin{aligned}
\!\!\mathop {\min }\limits_{{\bf{D}}^g,{\bf{A}}^g,{\bf{P}}^g}&\sum\limits_{g = 1}^{G - 1} \!\!\left\{ {\left\| {{{\bf{X}}^g} -{{\bf{H}}^g} {{\bf{D}}^g}{{\bf{A}}^g} - {{\bf{P}}^g}} \right\|_F^2 +   \gamma \left\| {{{\bf{P}}^g}} \right\|_F^2  }\right.\\
&\!\!\left. +{\left\| {{{\bf{Y}}^g} - {{\bf{H}}^{g \!+\! 1}}{{\bf{D}}^{g \!+\! 1}}{{\bf{A}}^g} - {{\bf{P}}^g}} \right\|_F^2  + \lambda_1 {{\left\| {{{\bf{A}}^g}} \right\|}_1}}\right\}  \\
\!\!\text{s.t.}~{\kern 4pt} || &{{\bf D}}^g(:,d) ||_2 \le 1,\forall  d \!\in\! \{ 1, \!\cdots\!,\!k\},\forall  g \!\in\! \{ 1, \cdots \!,G\},
\end{aligned}
\end{equation}
where $||{\bf A}^g||_1=\sum\nolimits_{i = 1}^{n^g} {||{{\bf a}}_{i^g}|{|_1}} $.

In Eq.~\eqref{eq3}, the optimization w.r.t. ${{\bf{P}}^g}$ has the closed-form solution by fixing  ${{\bf{D}}^g}$ and ${{\bf{A}}^g}$, where $g=1,2,\cdots,G-1$. By fixing ${{\bf{P}}^g}$, the optimization w.r.t. ${{\bf{D}}^g}$ and ${{\bf{A}}^g}$ becomes a joint sparse coding problem~\cite{yang2012coupled}. Such joint sparse coding problem can be solved in the concatenated feature space of aging layers $({{\bf{X}}^g} - {{\bf{P}}^g})$ and  $({{\bf{Y}}^g} - {{\bf{P}}^g})$ , but not in each feature space separately, namely we have
 \begin{equation} \label{eq400}
\begin{array}{l}
\mathop {\min }\limits_{{\bf{D}}^g,{\bf{A}}^g}\!\! \sum\limits_{g = 1}^{G - 1} \! \left\{\!\! {\left\|\! {\left[\!\!\! \!{\begin{array}{*{20}{c}}
			{{{\bf{X}}^g} \!-\! {{\bf{P}}^g}}\\
			{{{\bf{Y}}^g} \!-\! {{\bf{P}}^g}}
			\end{array}} \!\!\!\right] \!\!-\!\! \left[ {\begin{array}{*{20}{c}}
			{{{\bf{H}}^g}{{\bf{D}}^g}}\\
			{{{\bf{H}}^{g + 1}}{{\bf{D}}^{g + 1}}}
			\end{array}} \right]\!\!{{\bf{A}}^g}}\! \right\|_F^2 \!\!+\! {\lambda _1}{{\left\| {{{\bf{A}}^g}} \right\|}_1}} \!\!\right\}\\
{\rm{s}}{\rm{.t}}{\rm{.}}||{{\bf{D}}^g}(:,d)|{|_2} \le 1,\forall d \in \{ 1, \cdots ,k\} ,\forall g \in \{ 1, \cdots ,G\} ,
\end{array}
\end{equation}
which can be effectively solved by the SPAMS toolbox\footnote{http://spams-devel.gforge.inria.fr/}.

\subsection{Bi-level Dictionary Learning}
If the aging dictionaries ${\bf D}^1, \cdots, {\bf D}^G$ are learned by the coupled dictionary learning model, for an input face ${\bf x}^g$ belonging to the $g$-th age group, its sparse representation ${\bf a}^g$ and personalized layer ${\bf p}^g$ should be computed to generate its aging face aging face ${\bf \hat y}^{g}$ in the ($g+1$)-th age group.
According to Eq.~\eqref{eq3}, ${\bf a}^g$ and ${\bf p}^{g}$ should be calculated by solving the following problem,
\begin{equation} \label{eq511}
\begin{aligned}
\!\!\mathop {\min }\limits_{{\bf a}^g,{\bf p}^g}& \left\| {{{\bf{x}}^g} -{{\bf{H}}^g} {{\bf{D}}^g}{{\bf{a}}^g} - {{\bf{p}}^g}} \right\|_2^2 +   \gamma \left\| {{{\bf{p}}^g}} \right\|_2^2 \\
& +{\left\| {{{\bf{y}}^g} - {{\bf{H}}^{g \!+\! 1}}{{\bf{D}}^{g \!+\! 1}}{{\bf{a}}^g} - {{\bf{p}}^g}} \right\|_2^2  + \lambda_1 {{\left\| {{{\bf{a}}^g}} \right\|}_1}}\\
\end{aligned}.
\end{equation}

Unfortunately, ${\bf{y}}^g$ is unavailable for a new coming face ${\bf{x}}^g$ in  practice. Thus, there is no way to enforce the equivalence constraint on the  representations of  ${\bf{x}}^g$ and ${\bf{y}}^g$,  as  has  been  done  in  the  training phase of Eq.~\eqref{eq3}. To handle this problem, an intuitive way it to use the average face in the target age group to replace ${\bf y}^{g}$, as done in the preliminary work~\cite{shu2015personalized}. However, this solution has two drawbacks. First, since the average face and the target aging face have the different facial characteristics, the aging result is not satisfied. Second, to obtain the desired aging faces, we need to repeat the process of age progression synthesis several times to improve the aging performance, which is time-consuming.

To better address the above problem, let us revisit our primary goal again. We want to learn the aging dictionaries ${\bf D}^1, \cdots, {\bf D}^G$ to ensure that the sparse representations of two faces in each pair, as well as their personalized layers, are the same. If so, we can generate the aging face ${\bf \hat y}^{g}$ for an input face ${\bf x}^g$ by using the dictionary ${\bf D}^{g+1}$, the sparse representation ${\bf a}^g$ and the personalized layer ${\bf p}^g$, i.e., ${\bf \hat y}^{g}={{\bf{H}}^{g+1}} {{\bf{D}}^{g+1}}{{\bf a }^g} + {{\bf p}^g}$. Formally, an  ideal pair of every two neighboring aging  dictionaries  ${\bf D}^g$  and  ${\bf D}^{g+1}$
should satisfy the following two equations for all face pairs $\{{\bf x}_{i^g}, {\bf y}_{i^g}\}_{i^g=1}^{n^g}$ spanning the $g$-th and  ($g+1$)-th age groups,
\begin{equation} \label{eq61}
\begin{aligned}
\!\!\!\!\{ {\bf{a}}_{i^g}^*,{\bf{p}}_{i^g}^*\}  = \arg \mathop {\min }\limits_{{\bf{a}}_{i^g},{\bf{p}}_{i^g}} \sum\limits_{i^g=1}^{n^g} &{\left\{ {\left\| {{\bf{x}}_{i^g} - {{\bf{H}}^g}{{\bf{D}}^g}{\bf{a}}_{i^g} - {\bf{p}}_{i^g}} \right\|_2^2} \right.} \\
& \left. { + \gamma \left\| {{\bf{p}}_{i^g}} \right\|_2^2 + {\lambda _1}{{\left\| {{\bf{a}}_{i^g}} \right\|}_1}} \right\};
\end{aligned}
\end{equation}
\begin{equation} \label{eq62}
\!\!\{ {\bf a }_{i^g}^*,\!{\bf p }_{i^g}^*\}  \!=\! \arg \!\mathop {\min }\limits_{{\bf{a}}_{i^g},{\bf{p}}_{i^g}} \sum\limits_{i^g=1}^{n^g} {\!\left\| {{\bf{y}}_{i^g} \!-\! {{\bf{H}}^{g \!+\! 1}}{{\bf{D}}^{g \!+\! 1}}{\bf{a}}_{i^g} \!-\! {\bf{p}}_{i^g}} \right\|_2^2}.\!\!
\end{equation}

Based on the above analysis, we should first learn ${\bf {a}}^{g}$ and ${\bf{p}}^{g}$ by Eq.~\eqref{eq61}. And then ${\bf{D}}^{g}$ and ${\bf{D}}^{g+1}$ should be learned to satisfy  Eq.~\eqref{eq61} and Eq.~\eqref{eq62}. To this end, we can reformulate our objective function as
\begin{equation}\label{eq63}
\begin{array}{l}
\mathop {\min }\limits_{{{\bf{D}}^g},{{\bf{D}}^{g + 1}}} {\kern 1pt} {\kern 1pt} \sum\limits_{i^g = 1}^{n^g} {\left\{ {\left\| {{\bf{x}}_{i^g} - {{\bf{H}}^g}{{\bf{D}}^g}{\bf{a}}_{i^g} - {\bf{p}}_{i^g}} \right\|_2^2} \right.} \\
{\kern 1pt} {\kern 1pt} {\kern 1pt} {\kern 1pt} {\kern 1pt} {\kern 1pt} {\kern 1pt} {\kern 1pt} {\kern 1pt} {\kern 1pt} {\kern 1pt} {\kern 1pt} {\kern 1pt} {\kern 1pt} {\kern 1pt} {\kern 1pt} {\kern 1pt} {\kern 1pt} {\kern 1pt} {\kern 1pt} {\kern 1pt} {\kern 1pt} {\kern 1pt} {\kern 1pt} {\kern 1pt} {\kern 1pt} {\kern 1pt} {\kern 1pt} {\kern 1pt} {\kern 1pt} {\kern 1pt} {\kern 1pt} {\kern 1pt} {\kern 1pt} {\kern 1pt} {\kern 1pt} {\kern 1pt} {\kern 1pt} {\kern 1pt} \left. { + \left\| {{\bf{y}}_{i^g} - {{\bf{H}}^{g + 1}}{{\bf{D}}^{g + 1}}{\bf{a}}_{i^g} - {\bf{p}}_{i^g}} \right\|_2^2} \right\}\\
s.t. \!{\kern 1pt} {\kern 1pt}  \{ {\bf{a}}_{i^g},{\bf{p}}_{i^g}\}  \!=\! \arg \mathop {\min }\limits_{{\bf{a}}_{i^g},{\bf{p}}_{i^g}} \sum\limits_{i^g = 1}^{n^g} {\left\{ {\left\| {{\bf{x}}_{i^g} \!-\! {{\bf{H}}^g}{{\bf{D}}^g}{\bf{a}}_{i^g} \!-\! {\bf{p}}_{i^g}} \!\right\|_2^2} \right.} \\
{\kern 5pt} {\kern 6pt} {\kern 6pt} {\kern 1pt} {\kern 9pt} {\kern 7pt} {\kern 1pt} {\kern 1pt} {\kern 1pt} {\kern 1pt} {\kern 1pt} {\kern 1pt} {\kern 1pt} {\kern 1pt} {\kern 1pt} {\kern 1pt} {\kern 1pt} {\kern 1pt} {\kern 1pt} {\kern 1pt} {\kern 1pt} {\kern 1pt} {\kern 1pt} {\kern 1pt} {\kern 1pt} {\kern 1pt} {\kern 1pt} {\kern 1pt} {\kern 1pt} {\kern 1pt} {\kern 1pt} {\kern 1pt} {\kern 1pt} {\kern 1pt} {\kern 1pt} {\kern 1pt} {\kern 1pt} {\kern 1pt} {\kern 1pt} {\kern 1pt} {\kern 1pt} {\kern 1pt} {\kern 1pt} {\kern 1pt} {\kern 1pt} {\kern 1pt} {\kern 1pt} {\kern 1pt} {\kern 1pt} {\kern 1pt} {\kern 1pt} {\kern 1pt} {\kern 1pt} {\kern 1pt} {\kern 1pt} {\kern 1pt} {\kern 1pt} {\kern 1pt} {\kern 1pt} {\kern 1pt} {\kern 1pt} {\kern 1pt} {\kern 1pt} {\kern 1pt} {\kern 1pt} {\kern 1pt} {\kern 1pt} {\kern 1pt} {\kern 1pt} {\kern 1pt} {\kern 1pt} {\kern 1pt} {\kern 1pt} {\kern 1pt} {\kern 1pt} {\kern 1pt} {\kern 1pt} {\kern 1pt} {\kern 1pt} {\kern 1pt} {\kern 1pt} {\kern 1pt} {\kern 1pt} {\kern 1pt} {\kern 1pt} {\kern 1pt} {\kern 1pt} {\kern 1pt} {\kern 1pt} {\kern 1pt} {\kern 1pt} {\kern 1pt} {\kern 1pt} {\kern 1pt} {\kern 1pt} {\kern 1pt} {\kern 1pt} {\kern 1pt} {\kern 1pt} {\kern 1pt} {\kern 1pt} {\kern 1pt} {\kern 1pt} {\kern 1pt} {\kern 1pt} \left. { + \gamma \left\| {{\bf{p}}_{i^g}} \right\|_2^2 \!+\! {\lambda _1}{{\left\| {{\bf{a}}_{i^g}} \right\|}_1}} \right\}\\
{\kern 1pt} {\kern 1pt} {\kern 1pt} {\kern 1pt} {\kern 1pt} {\kern 1pt} {\kern 1pt} {\kern 1pt} {\kern 1pt} {\kern 1pt} {\kern 1pt} {\kern 1pt} {\kern 1pt} {\kern 1pt} {\kern 1pt} {\left\| {{{\bf{D}}^c}(:,d)} \right\|_2} \le 1,{\kern 1pt} {\kern 1pt} {\kern 1pt} \forall  d \!\in\! \{ 1, \cdots ,k\},{\kern 1pt} {\kern 1pt} {\kern 1pt} c = \{ g,g + 1\} ,
\end{array}
\end{equation}
which is a bi-level optimization problem~\cite{yang2012bilevel}.

Rewrite Eq.~\eqref{eq63} in the matrix form after some algebratic steps, we can obtain a novel bi-level dictionary learning model for all face pairs $\{{\bf x}_{i^g}, {\bf y}_{i^g}\}_{i^g=1}^{n^g}$ spanning the $g$-th and  ($g+1$)-th age groups ($g=1,2,
\cdots,G-1$):
\begin{equation}\label{eq5.1}
\begin{array}{l}
\mathop {\min }\limits_{{{\bf{D}}^g},{{\bf{D}}^{g + 1}}} J({{\bf{D}}^g},{{\bf{D}}^{g + 1}})\\
s.t.{\kern 1pt} {\kern 1pt} {\kern 1pt} {\kern 1pt} {\kern 1pt} {\kern 1pt} {{\bf{A}}^g} = \arg \mathop {\min }\limits_{{{\bf{Z}}^g}} \left\| {{{\bf{X}}^g} - {{\bf{H}}^g}{{\bf{D}}^g}{{\bf{Z}}^g} - {{\bf{P}}^g}} \right\|_F^2 + \lambda_1 {\left\| {{{\bf{Z}}^g}} \right\|_1}\\{\kern 80pt}+\lambda_2{\left\| {{{\bf{Z}}^g}} \right\|_F^2}\\
{\kern 1pt} {\kern 1pt} {\kern 1pt} {\kern 1pt} {\kern 1pt} {\kern 1pt} {\kern 1pt} {\kern 1pt} {\kern 1pt} {\kern 1pt} {\kern 1pt} {\kern 1pt} {\kern 1pt} {\kern 1pt} {\kern 1pt} {\kern 1pt} {\kern 1pt} {\kern 1pt} {\kern 1pt} {{\bf{P}}^g} = \arg \mathop {\min }\limits_{{{\bf{Q}}^g}} \left\| {{{\bf{X}}^g} - {{\bf{H}}^g}{{\bf{D}}^g}{{\bf{Z}}^g} - {{\bf{Q}}^g}} \right\|_F^2 + \gamma \left\| {{{\bf{Q}}^g}} \right\|_F^2\\
{\kern 1pt} {\kern 1pt} {\kern 1pt} {\kern 1pt} {\kern 1pt} {\kern 1pt} {\kern 1pt} {\kern 1pt} {\kern 1pt} {\kern 1pt} {\kern 1pt} {\kern 1pt} {\kern 1pt} {\kern 1pt} {\kern 1pt} {\kern 1pt} {\kern 1pt} {\kern 1pt} {\kern 1pt} \!{\left\| {{{\bf{D}}^c}(:,d)} \right\|_2} \!\le\! 1,{\kern 1pt} {\kern 1pt} {\kern 1pt} \forall  d \!\in\! \{ 1, \cdots\! ,k\},{\kern 1pt} {\kern 1pt} {\kern 1pt} {\text{and}}{\kern 1pt} {\kern 1pt} c =\! \{ g,g + \!1\},
\end{array}
\end{equation}
where $J ({{\bf{D}}^g},{{\bf{D}}^{g + 1}})= \left\| {{{\bf{X}}^g} - {{\bf{H}}^g}{{\bf{D}}^g}{{\bf{A}}^g} - {{\bf{P}}^g}} \right\|_F^2 + \left\| {{{\bf{Y}}^g} - {{\bf{H}}^{g + 1}}{{\bf{D}}^{g + 1}}{{\bf{A}}^g} - {{\bf{P}}^g}} \right\|_F^2{\kern 1pt}$, ${\bf Q}^g=[{\bf q}_{1^g},\cdots,{\bf q}_{n^g}]\in \mathbb{R}^{f\times n^g}$, ${\bf Z}^g=[{\bf z}_{1^g},\cdots,{\bf z}_{n^g}]\in \mathbb{R}^{k \times n^g}$, and $\lambda_2$ controls the regularization.

According to the new formulation in Eq.~\eqref{eq5.1}, for a new coming face ${\bf x}^g$, the calculations of its sparse representation ${\bf a}^g$ and personalized layer ${\bf p}^g$ are independent with the unavailable ${\bf y}^g$, while they only depend on the available face ${\bf x}^g$.


\subsection{Optimization Procedure}
\label{OP}
The objective function in Eq.~\eqref{eq5.1} is convex w.r.t. ${\bf D}^{g+1}$ and ${\bf D}^{g}$ separately, which can be iteratively solved through two alternating sub-procedures of optimization. Specifically, we fix the other variables when updating one variable.

\subsubsection{Updating ${\bf D}^{g+1}$}
We update ${\bf D}^{g+1}$ by fixing ${\bf D}^{g}$, and then the objective function in Eq.~\eqref{eq5.1} becomes
\begin{equation} \label{eq3.1}
\begin{array}{l}
\mathop {\min }\limits_{{\bf D}^{g+1}} \left\| {{{\bf{Y}}^g} - {{\bf{H}}^{g + 1}}{{\bf{D}}^{g + 1}}{{\bf{A}}^g} - {{\bf{P}}^g}} \right\|_F^2{\kern 1pt}\\  
s.t.{\kern 1pt} {\kern 1pt} {\kern 1pt} {\kern 1pt} {\kern 1pt} {\kern 1pt} {{\bf{A}}^g} = \arg \mathop {\min }\limits_{{{\bf{Z}}^g}} \left\| {{{\bf{X}}^g} - {{\bf{H}}^g}{{\bf{D}}^g}{{\bf{Z}}^g} - {{\bf{P}}^g}} \right\|_F^2 + \lambda_1 {\left\| {{{\bf{Z}}^g}} \right\|_1}\\{\kern 80pt}+\lambda_2{\left\| {{{\bf{Z}}^g}} \right\|_F^2}\\
{\kern 1pt} {\kern 1pt} {\kern 1pt} {\kern 1pt} {\kern 1pt} {\kern 1pt} {\kern 1pt} {\kern 1pt} {\kern 1pt} {\kern 1pt} {\kern 1pt} {\kern 1pt} {\kern 1pt} {\kern 1pt} {\kern 1pt} {\kern 1pt} {\kern 1pt} {\kern 1pt} {\kern 1pt} {{\bf{P}}^g} = \arg \mathop {\min }\limits_{{{\bf{Q}}^g}} \left\| {{{\bf{X}}^g} - {{\bf{H}}^g}{{\bf{D}}^g}{{\bf{A}}^g} - {{\bf{Q}}^g}} \right\|_F^2 + \gamma \left\| {{{\bf{Q}}^g}} \right\|_F^2\\
{\kern 1pt} {\kern 1pt} {\kern 1pt} {\kern 1pt} {\kern 1pt} {\kern 1pt} {\kern 1pt} {\kern 1pt} {\kern 1pt} {\kern 1pt} {\kern 1pt} {\kern 1pt} {\kern 1pt} {\kern 1pt} {\kern 1pt} {\kern 1pt} {\kern 1pt} {\kern 1pt} {\kern 1pt} {\left\| {{{\bf{D}}^{g+1}}(:,d)} \right\|_2} \le 1,{\kern 1pt} {\kern 1pt} {\kern 1pt} \forall  d \!\in\! \{ 1, \cdots ,k\}.
\end{array}
\end{equation}
First, we iteratively update ${\bf A}^g$ and ${\bf P}^g$, wherein the $lasso$ problem can be solved by the SPAMS toolbox. And then, the problem in Eq.~\eqref{eq3.1} becomes a Quadratically Constrained Quadratic Program (QCQP) that can be solved by the CVX toolbox\footnote{http://cvxr.com/cvx/}.

\begin{algorithm}[t]
	\scriptsize{
		\renewcommand{\algorithmicrequire}{\textbf{Input:}}
		\renewcommand\algorithmicensure {\textbf{Output:} }
		\caption{Bi-level Dictionary Learning (Offline)}
		\small
		\label{alg1}
		\begin{algorithmic}[1]
			\REQUIRE
			{$\{{\bf X}^g,{\bf Y}^g\}_{g=1}^{G-1}$, ${\bf H}^g$ ($g=1,\cdots,G-1$), $\lambda_1$, $\lambda_2$, and $\gamma$.}
			\renewcommand{\algorithmicrequire}{\textbf{Initialization:}}
			\REQUIRE
			${\bf D}=[{\bf D}^1,{\bf D}^2,\cdots,{\bf D}^G]$ by Coupled Dictionary Learning in ~\cite{shu2015personalized}, $n_0 \leftarrow1$, and
			$iter\leftarrow0$.
			\FOR{$g = 1,2,\cdots, G-1$}
			\REPEAT
			\FOR{$i^g = 1,2,\cdots, n^g$}
			\STATE Update index set $\Omega$ based on Definition~\ref{d1};
			\STATE Compute gradient
			$\Delta  = \partial J/\partial {{\bf{D}}^g}$ with Eq.~\eqref{eq71};
			\STATE Update ${{\bf{D}}^g}={{\bf{D}}^g}-\eta(n_0)\cdot \Delta$\footnotemark[1];
			\STATE Project each atom of ${{\bf{D}}^g}$ onto the unit ball;
			\STATE $n_0 \leftarrow n_0+1$.
			\ENDFOR
			\STATE Update ${\bf D}^{g+1}$ with Eq.~\eqref{eq3}.
			\STATE $iter \leftarrow iter+1$.
			\UNTIL{convergence.}
			\ENDFOR		
			\ENSURE{${\bf D}=[{\bf D}^1,{\bf D}^2,\cdots,{\bf D}^G]$.}
		\end{algorithmic}
	}
	\footnotemark[1]{ here, $\eta(n_0)$ shrinks in the rate of $1/n_0$}.
\end{algorithm}

\subsubsection{Updating ${\bf D}^g$}
When updating ${\bf D}^g$, we fix ${\bf D}^{g+1}$, and then Eq.~\eqref{eq5.1} becomes a bi-level optimization problem. Similar to~\cite{yang2012bilevel,yang2012coupled}, we solve this bi-level optimization problem based on the first-order projected stochastic gradient descent. 
For brevity, we simplify the subscripts of ${\bf x}_{i^g}$, ${\bf x}_{i^g}$, ${\bf a}_{i^g}$ and ${\bf p}_{i^g}$, i.e., ${\bf x}{^g}$, ${\bf y}^g$, ${\bf a}^{g}$ and ${\bf p}^{g}$. For ${\bf x}^g$,
we compute the gradient of $J$ with respect to ${\bf D}^g$
\begin{equation}
\begin{aligned}
\label{eq71}
\!\!\frac{{\partial J}}{{\partial {{\bf{D}}^g}}}{\kern 1pt} {\kern 1pt}  \!\!\!=\!\!  &\sum\limits_{j \in \Omega } \!\!{\frac{{\partial (\!{J_x} \!\!+\!\! {J_y}\!)}}{{\partial \tilde a_j^g}}\!\frac{{\partial \tilde a_j^g}}{{\partial {{\bf{D}}^g}}}} \! + \!\!\sum\limits_{i = 1}^f \!\!{\frac{{\partial (\!{J_x}\! +\! {J_y}\!)}}{{\partial p_i^g}}\!\frac{{\partial p_i^g}}{{\partial {{\bf{D}}^g}}}}  \!\!+\!\! \frac{{\partial {J_x}}}{{\partial {{\bf{D}}^g}}}\!,\!
\end{aligned}
\end{equation}
where  $J_x = \left\| {{{\bf{X}}^g} - {{\bf{H}}^g}{{\bf{D}}^g}{{\bf{A}}^g} - {{\bf{P}}^g}} \right\|_F^2$, and $J_y= \left\| {{{\bf{Y}}^g} - {{\bf{H}}^{g + 1}}{{\bf{D}}^{g + 1}}{{\bf{A}}^g} - {{\bf{P}}^g}} \right\|_F^2{\kern 1pt}$. $\Omega$ denotes the
index set for $j$, and is defined as follows,
\begin{myDef}
	\label{d1}
Let ${\bf w}^g={\bf x}^g-{\bf p}^g$, we solve the following $lasso$ problem to obtain the optimizing ${\bf a}^g$:
\begin{equation}
	\label{eq5.a}
	\min_{{\bf a}^g}||{{\bf{w}}^g} - {\bf{H}}^g{{{\bf{D}}}^g}{{{\bf{a}}}^g}||_2^2 + \lambda ||{{{\bf{a}}}^g}|{|_1}+ \lambda_2 ||{{{\bf{a}}}^g}|{|_2^2}.
\end{equation}
If we obtain ${\bf a}^g$ by Eq.~\eqref{eq5.a}, we can define $\Omega  = \{ j||{a^g_j}| > {0^ + }\} $ is the index set of nonzero elements of ${\bf a}^g$.
\end{myDef}


In Eq.~\eqref{eq71}, it is easy to find that
\begin{equation}
\label{eq81}
\frac{{\partial {J_x}}}{{\partial {{\bf{D}}^g}}} =  - 2{({{\bf{H}}^g})^T}{{\bf{u}}^g}{({{\bf{a}}^g})^T} + 2{({{\bf{H}}^g})^T}{{\bf{H}}^g}{{\bf{D}}^g}{{\bf{a}}^g}{({{\bf{a}}^g})^T},
\end{equation}
where ${\bf{u}}^g={{\bf x}}^g-{{\bf p}}^g$. The detailed deduction of $\sum\limits_{j \in \Omega } {\frac{{\partial ({J_x} + {J_y})}}{{\partial \tilde a_j^g}}\frac{{\partial \tilde a_j^g}}{{\partial {{\bf{D}}^g}}}}$  and $\sum\limits_{i = 1}^f {\frac{{\partial ({J_x} + {J_y})}}{{\partial p_i^g}}\frac{{\partial p_i^g}}{{\partial {{\bf{D}}^g}}}}$ in Eq.~\eqref{eq71}  can be found in Appendix~A.1 and~A.2 of the supplemental material, respectively.
Finally, we can obtain the updating way of ${\bf D}^{g}$, as follows,
\begin{equation}
\label{eq82}
{{\bf{D}}^g}={{\bf{D}}^g}-\eta \cdot \frac{{\partial J}}{{\partial {{\bf{D}}^g}}},
\end{equation}
where $\eta$ is the step size.

\begin{figure}[t]
	\centering
	\includegraphics[scale=0.38]{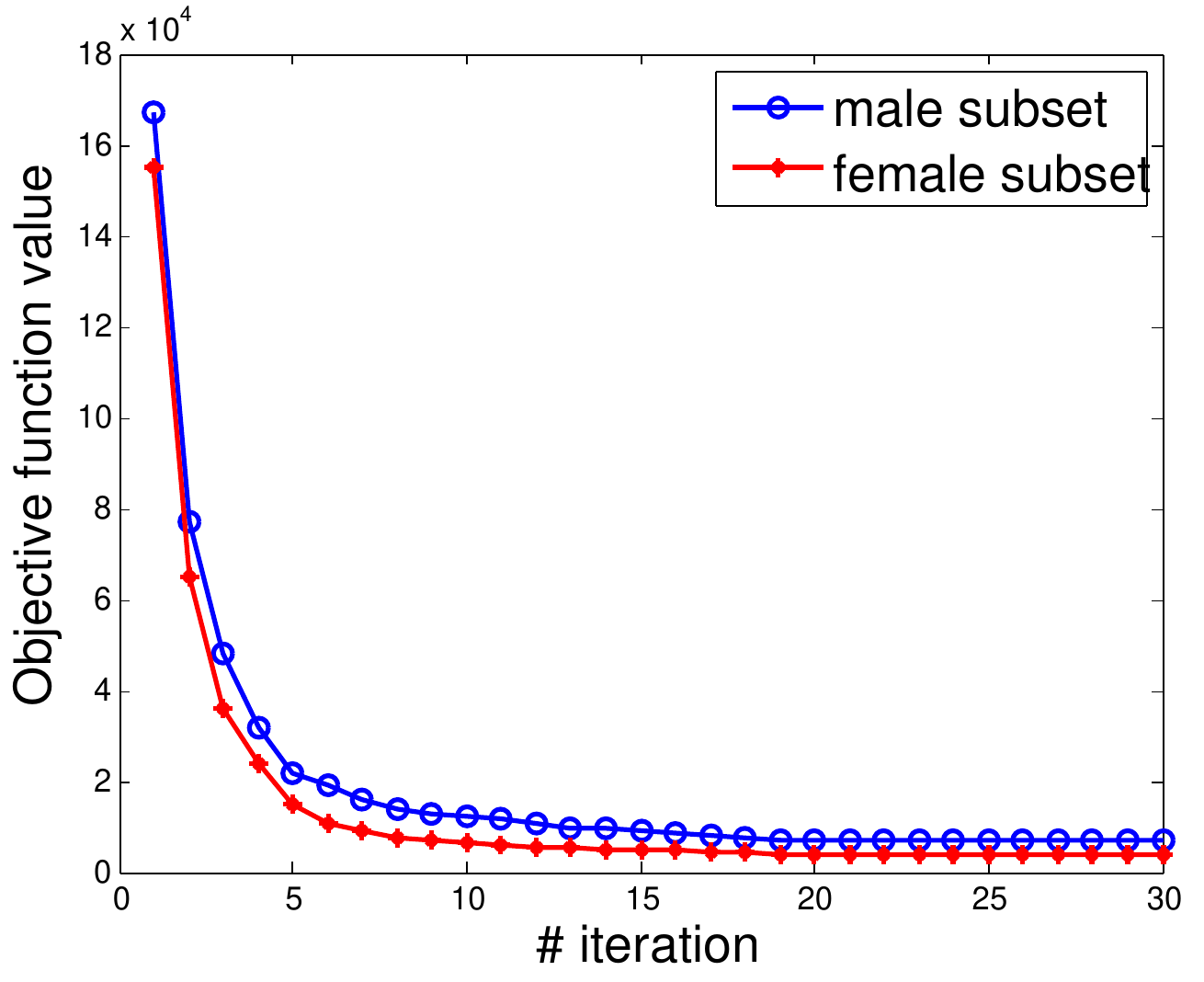}
	\caption{Convergence curves of the optimization procedure for solving ${\bf D}^{1}$ and ${\bf D}^{2}$.}
	\label{fig33}
\end{figure}

The proposed bi-level dictionary learning algorithm is summarized in Algorithm~\ref{alg1}. The convergence criterion is that the iteration steps shall end when the relative cost of objective function is smaller than a pre-defined threshold. The convergence curves for solving ${\bf D}^{1}$ and ${\bf D}^{2}$ on male and female sub datasets respectively are shown in Figure~\ref{fig33}. We found that the algorithm achieves convergence after about $30$ iterations.

\begin{algorithm}[t]
	\scriptsize{
		\renewcommand{\algorithmicrequire}{\textbf{Input:}}
		\renewcommand\algorithmicensure {\textbf{Output:} }
		\caption{Age Progression Synthesis (Online)}
		\small
		\label{alg2}
		\begin{algorithmic}[1]
			\REQUIRE
			{${\bf D}=[{\bf D}^1,\cdots,{\bf D}^c,\cdots,{\bf D}^G]$, input image ${\bf x}^g$ in the $g$-th age group, $\{{\bf D}^g,{\bf D}^{g+1}\}_{g=1}^{G-1}$, $\lambda_1$, $\lambda_2$ and $\gamma$}.
			\FOR{$t = 1,\cdots, G-g$}
			\STATE Solve ${{\bf a}^{g+t-1}}^{*}$ and ${{\bf p}^{g+t-1}}^{*}$ with Eq.~\eqref{eq61}
			\STATE Obtain ${\bf \hat y}^{g+t-1}$:\\
			$ {{{\bf \hat y}^{g+t-1}} = {{\bf{H}}^{g + t}}{{\bf{D}}^{g + t}}{{\bf a}^{g+t-1}}^{*} + {{\bf a}^{g+t-1}}^{*} }; $
			\ENDFOR
			\ENSURE{$\{{\bf \hat y}^{g}, {\bf \hat y }^{g+1},\cdots,{\bf \hat y}^{G-1}\}$.}
		\end{algorithmic}
	}
\end{algorithm}

\subsection{Age Progression Synthesis}
\label{APS}
After learning all aging dictionaries ${{\bf D}^1,{\bf D}^2,\cdots, {\bf D}^G}$, for a given face ${\bf x}^g$ belonging to the $g$-th age group\footnote{Here, its age range and gender are estimated by an age estimator and a gender recognition system in~\cite{li2015shape}, respectively.}, we can generate its aging face sequence $\{{\bf \hat y}^{g},...,{\bf \hat y}^{G-1}\}$ step by step, from the current age to the target age. Specifically, we first generate the aging face ${\bf \hat y}^{g}$ in the next age group (i.e., the ($g+1$)-th age group) by the corresponding aging dictionary with a sparse representation ${\bf  a}^{g*}$, as well as a personalized layer ${\bf  p}^{g*}$. Here, ${\bf  a}^{g*}$ and ${\bf p}^{g*}$ are calculated by Eq.~\eqref{eq61} in an alternately iterative way. The iteration steps of optimization in Eq.~\eqref{eq61}
shall end when the relative cost of objective function is smaller than a pre-defined threshold.

When we have obtained ${\bf a}^{g*}$ and ${\bf p}^{g*}$, the aging face ${\bf \hat y}^{g}$ in the next age group (i.e., the ($g+1$)-th age group) can be generated by the following equation:
\begin{equation}
\label{eq83.1}
{{{\bf \hat y }^{g}} = {{\bf{H}}^{g + 1}}{{\bf{D}}^{g + 1}}{{\bf a}^{g*}} + {{\bf p}^{g*}}}.
\end{equation}
After that, taking this new aging face ${\bf \hat y}^{g}$
in the  ($g+1$)-th age group as the input of age synthesis for the  ($g+2$)-th age group. We repeat this process until all
aging faces ${\bf \hat y}^{g}, {\bf \hat y}^{g+1}, ...,{\bf \hat y}^{G-1}$ are generated.

In the above process, we do not need to calculate the average face for each input face, as done in the preliminary work~\cite{shu2015personalized}, and then do not need to repeat the process of age progression synthesis. Therefore, we can save much time compared with the preliminary work~\cite{shu2015personalized}.

\begin{figure*}[!th]
	\centering
	\includegraphics[scale=0.20]{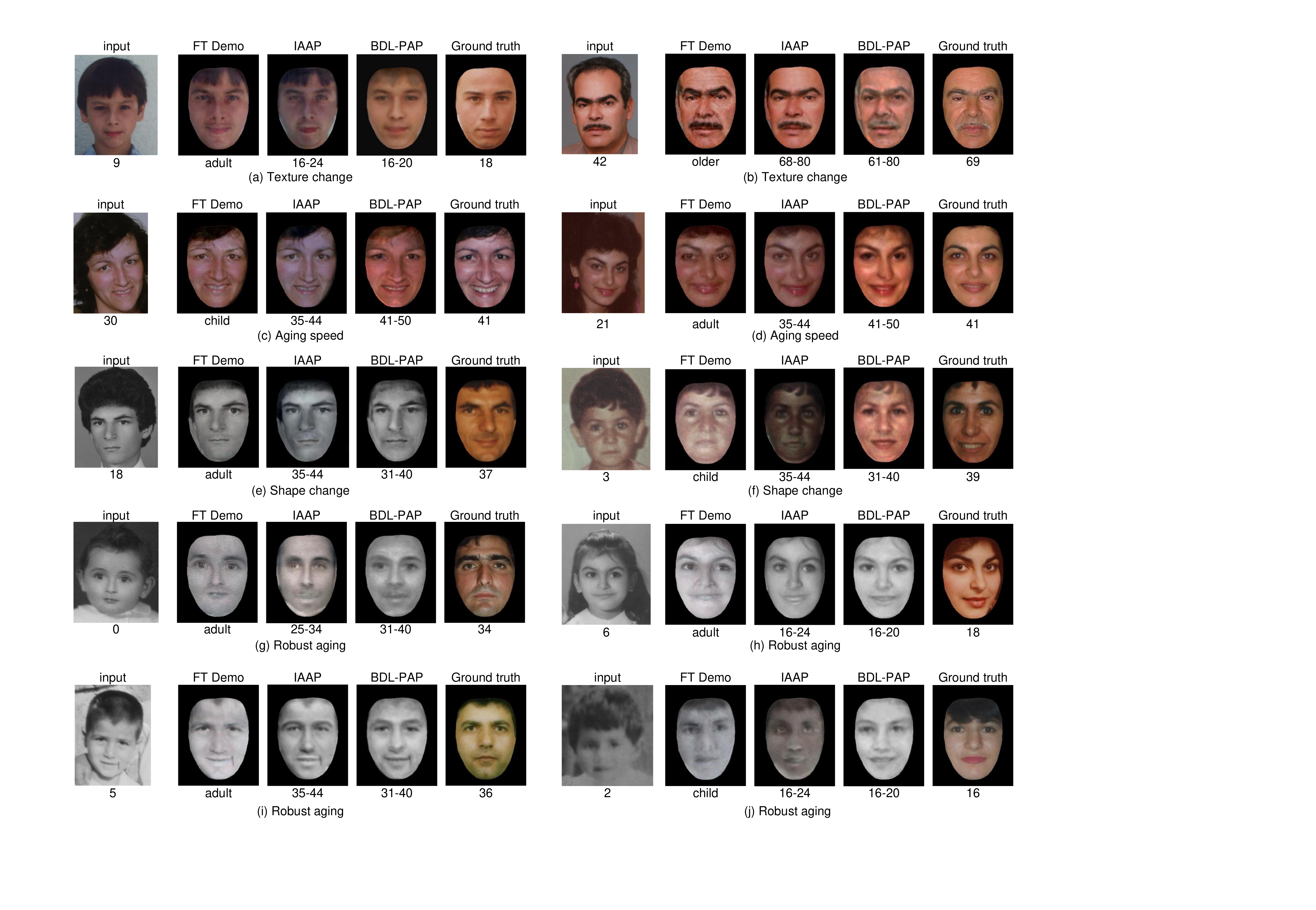}
	\caption{The comparisons with ground truth and other methods. Each group includes an input face, a ground truth and three aging results generated by different methods. The number or word under each face photo represents the age range (e.g., 61-80) or the age period (e.g., older). For convenience of comparison, black background has been added to each face photo. Better view in color.}
	\label{fig4}
\end{figure*}

\section{Experiments}
\label{Exe}
\subsection{Experimental Settings}
\label{ID}
{\bf Data collection.} To train the high-quality aging dictionary, it is crucial to collect sufficient and dense short-time face aging pairs. We download a large number of face photos covering
different ages of the same persons from Google and Bing image searches, social media websites(by some content and context links~\cite{qi2012exploring}), and other two available face aging datasets, CACD~\cite{chen2014cross} and MORPH~\cite{ricanek2006morph}.  Both CACD and MORPH contain quite a number of  short-term intra-person photos. Since face images from the Internet and CACD dataset are mostly ``in the wild", we select the photos with approximately frontal faces ($- {15^ \circ }$ to ${15^ \circ}$) and relatively natural illumination and expressions. For all face images, face alignment~\cite{viola2001rapid} is implemented to obtain aligned faces, which are cropped into the size $123\times 98\times 3$. To boost the aging relationship between the neighboring aging dictionaries, we employ Collection Flow~\cite{kemelmacher2012collection} to correct all the faces into the common neutral expression. We group all images (the age from 0 to 80) into $9$ age groups (i.e., $G=9$): 0-5, 6-10, 11-15, 16-20, 21-30, 31-40, 41-50, 51-60, and 61-80 of two genders, and find that no person has aging faces covering all aging groups. Actually, the aging faces of most persons fall into only one or two age groups (i.e., most persons have face photos spanning no more than 20 years).  Therefore, we further select those intra-person face photos which densely fall into two neighboring age groups. Finally, there are 3,200 intra-person face pairs for training (1,600 pairs for males, and 1,600 pairs for females)\footnote{Dataset is released at http://imag.njust.edu.cn/FAD.html}. Every two neighboring age groups for one gender share 400 face aging pairs of the same persons. Since male and female have different aging characteristics, we train aging dictionaries for male and female, respectively.

{\bf PCA projection.} We stack $s$ images in the $g$-th age group as columns of a data matrix ${\bf M}^g \!\in\! \mathbb{R}^{f\times s}$, where $s=400$ and $f=123\times 98\times 3=36,162$ in experiments. The SVD of ${\bf M}^g$ is ${{{\bf M}^g}} = {{{\bf U}^g}}{{{\bf S}^g}}({\bf V}^g)^T$. We define the projected matrix ${\bf H}^g={\bf U}^g(:,1\!:\!m)\in\mathbb{R}^{f \times m}$, where ${\bf U}^g(:,1\!:\!m)$ is truncated to the rank~=~$m$, and $m=2,000$ in experiments.

{\bf Parameter setting.} The parameters $\lambda_1$, $\lambda_2$ and $\gamma$ in Eq.~\eqref{eq3} are empirically set as $\lambda_1=0.01$, $\lambda_2=0.001$ and $\gamma=0.1$. The number of bases of each aging dictionary ${\bf D}^g$ is set as $k=80$.


{\bf Aging evaluation.}
We adopt three strategies to comprehensively evaluate the proposed age progression method. First, we qualitatively evaluate the proposed method on FG-NET~\cite{fgnet}. We show the age progression for every photo in FG-NET, and do the qualitative comparison with the corresponding ground truth (available older photos) for each person. For reference, we also reproduce some aging results of other representative methods. Second, we conduct user study to test the aging faces of the proposed method compared with the prior works which reported their best aging results. The proposed method uses the same inputs as in these prior works. Third, we compare the proposed method by the cross-age face verification~\cite{gong2013hidden,wu2012age}. Cross-age face recognition~\cite{chen2014cross,yadav2013bacteria} and cross-age face verification are challenging in extreme facial analysis scenarios due to the age gap, which is similar to the semantic gap between the image and tag in the field of computer vision~\cite{qi2016joint,qi2009two-dimensional}. A straightforward way for cross-age facial analysis is to use the aging synthesis to reduce the age gap. Specifically, we can synthesize all the faces to their aging faces within the same age range, and then implement the face verification algorithm. In turn, we can also use the face verification to validate whether the intra-person pair of aging face and ground truth face (without age gap) is more similar than the original intra-person face pair with the age gap.

\subsection{Qualitative Comparison with Ground Truth}
Since FG-NET provides the ground-truth aging faces, we compare the proposed BDL-PAP method with an online Face Transformer demo (i.e., FT Demo\footnote{http://cherry.dcs.aber.ac.uk/Transformer/}), and the representative Illumination-Aware Age
Progression (IAAP) method~\cite{kemelmacher2014illumination} on this dataset. FT Demo requires manual location of facial features, while IAAP uses common aging characteristics of average faces for the age progression of all input faces.

Some aging results generated by these three compared methods are given in Figure~\ref{fig4}, covering from baby/childhood/teenager (input) to adult/agedness (output), as well as from adult (input) to agedness (output). By comparing with ground truth, we can see that the aging results generated by the proposed BDL-PAP look more like the ground truth faces than the aging results of other two methods. In particular, the proposed BDL-PAP can generate personalized aging faces for different individual inputs. In term of texture change, the aging face of BDL-PAP in Figure~\ref{fig4}(a) has no mustache that is closer to ground truth, while the aging face of BDL-PAP in Figure~\ref{fig4}(b) has white mustache that is closer to ground truth; In term of shape change, the aging faces of BDL-PAP in Figure~\ref{fig4}(e)(f) have more approximate facial outline to the ground truth; In term of aging speed, the faces of FT Demo and IAAP in Figure~\ref{fig4}(d) are aging more slowly, while one of FT Demo in Figure~\ref{fig4}(c) is faster. Overall, the age speed of IAAP is slower than ground truth since IAAP is based on smoothed average faces, which maybe loses some facial texture details, such as freckle, nevus, aging spots, etc. FT Demo performs the worst, especially in shape change. Our aging results in Figure~\ref{fig4} are more similar to the ground truth, which means BDL-PAP can preserve much more personalized results. Moreover, as shown in Figure~\ref{fig4}(g)(h), the aging results of BDL-PAP are more robust than other methods for the input faces with noise or low resolution.

\subsection{Quantitative Comparison with Prior Works}

Some prior works related to the age progression have posted their best face aging results with input faces at different ages, including
\cite{suo2010compositional}, \cite{scherbaum2007prediction}, \cite{ramanathan2009age}, \cite{park2008face}, \cite{patterson2007comparison}, \cite{liang2007age}, \cite{liang2011multi}, \cite{shen2011exemplar}, \cite{sethuram2010hierarchical}, \cite{wang2006age}, \cite{wang2016recurrent} and \cite{kemelmacher2014illumination}. There are 261 aging results with 87 input faces in total. The proposed BDL-PAP for each input face is implemented to generate the corresponding aging faces at the same ages (ranges) of the posted results.

\begin{figure*}[!th]
	\centering
	\includegraphics[scale=0.24]{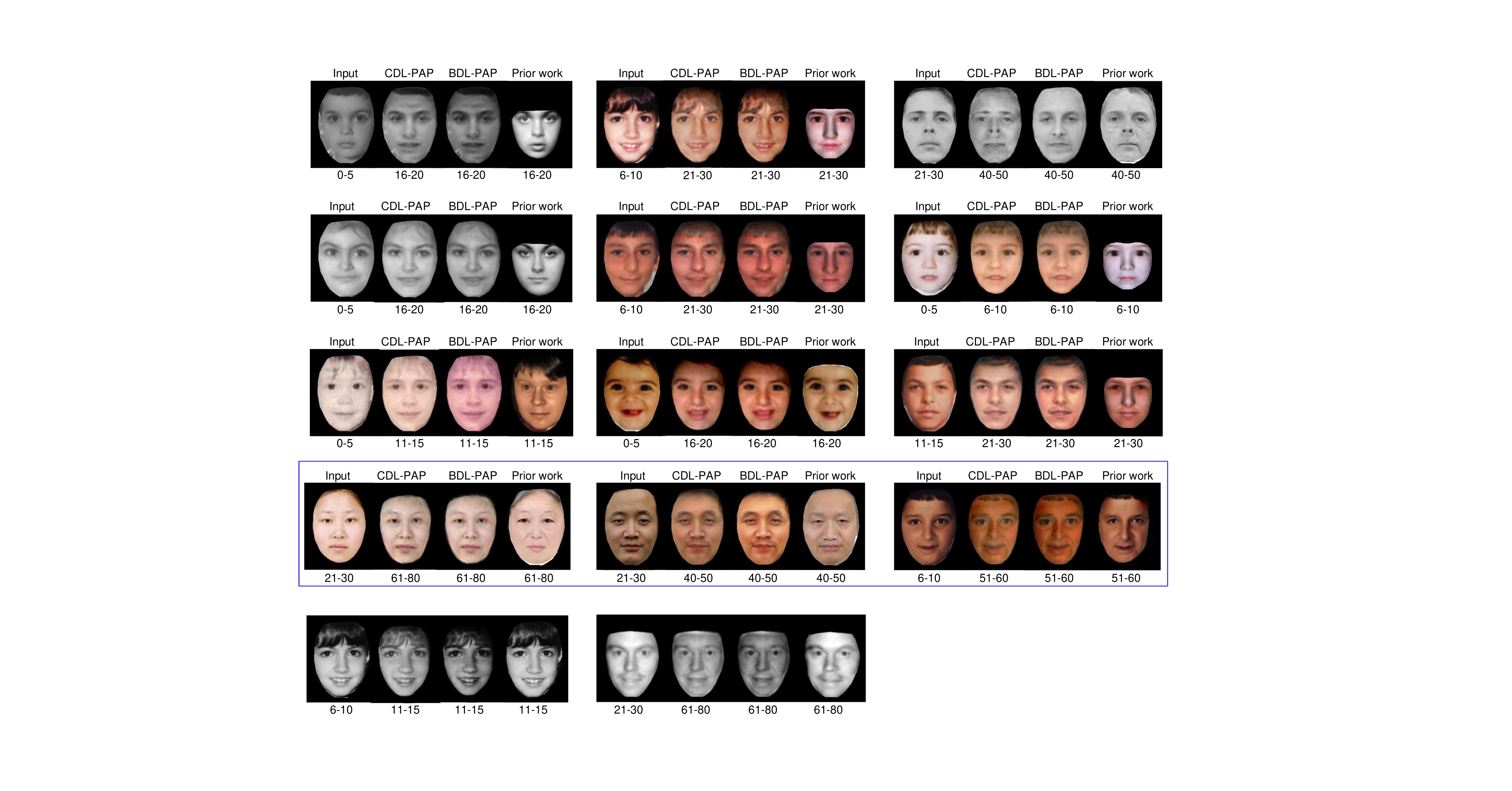}
	\caption{The comparisons with prior works. Each comparison group includes an input face and three aging results of CDL-PAP, BDL-PAP and prior work. The number under the face photo is the age range. Some worse aging results of BDL-PAP are enclosed by blue box. For convenience of comparison, black background has been added to each face photo.}
	\label{fig5}
\end{figure*}

We conduct user study to compare the aging results of the proposed BDL-PAP with the posted aging results in the prior works, as well as the aging results generated by Coupled Dictionary Learning based Personalized Age Progression (CDL-PAP) in the preliminary work~\cite{shu2015personalized}.
To avoid bias as much as possible, we invite 50 users covering a wide age range and from all walks of life. For each comp arison group including an input face, and three aging results in a random order, all the 50 users are asked to answer the question: which aging face is the best in terms of {\em Personality} and {\em Reliability}. {\em Reliability} means the aging face should be natural and authentic at the synthetic age, while {\em Personality} means the aging faces for different inputs should be identity-preserved and diverse. 
User can choose one from the three aging results as the best, and choose ``None" if she/he thinks all the three aging results are unsatisfied. There are 50 ratings for each comparison, 261 comparison groups, and then 13,050 ratings in total. The voting results are as follows: 36.5\% for BDL-PAP best; 34.8\% for CDL-PAP best; 26.7\% for prior works best; and 2.0\% for ``none is satisfied". 
We show some comparison groups for voting in Figure \ref{fig5}. Overall, for the input face of a person in any age range, BDL-PAP, CDL-PAP and these prior works can generate an authentic and reliable aging face of any older-age range. In particular, for different inputs, aging faces rendered by BDL-PAP and CDL-PAP have more personalized aging characteristics, which further improve the appealing visual sense. For example in Figure \ref{fig5}, the aging faces of BDL-PAP in the same age range in the first and the second group of the first column have different aging speeds: the former is faster than the latter; the aging faces of prior works with different inputs in the first and second groups of the third column are similar, while the aging results of BDL-PAP and CDL-PAP are more diverse for different individual inputs. Moreover, we can see that the aging results of BDL-PAP are comparable with the aging results of CDL-PAP. Specifically, BDL-PAP can synthesize the more sharp and high-definition aging faces than CDL-PAP.


\begin{figure}[t]
	\centering
	\includegraphics[scale=0.50]{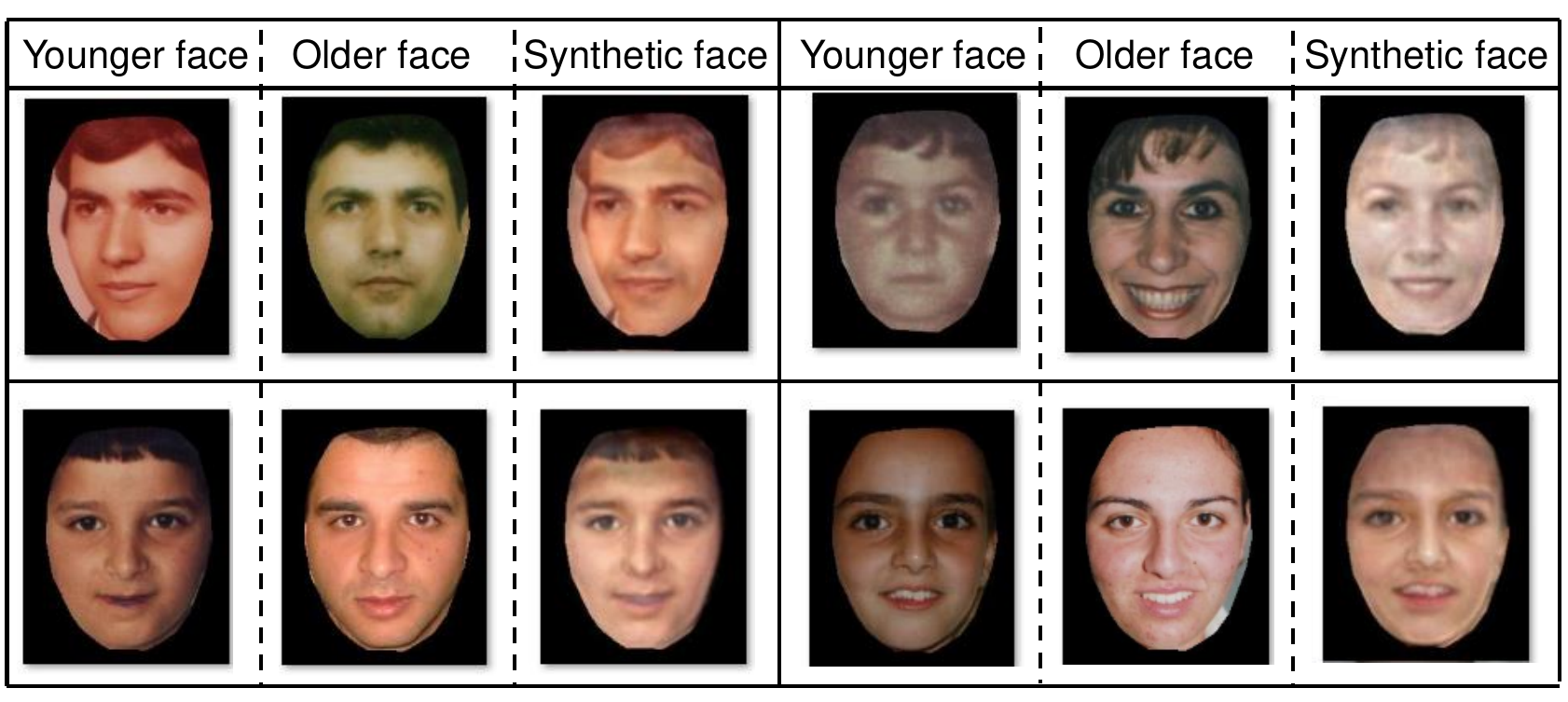}
	\caption{The comparisons of original face pairs and the synthetic pairs by BDL-PAP. The face images in each solid-line box are the same person. Original pair consists of younger face and older face, while synthetic pair consists of synthetic face and older face.}
	\label{fig_show_pair}
\end{figure}

\subsection{Evaluation on Cross-Age Face Verification}
\begin{figure}[!t]
	
	\centering
	\subfigure[Pair setting.]{
		\includegraphics[scale=0.20]{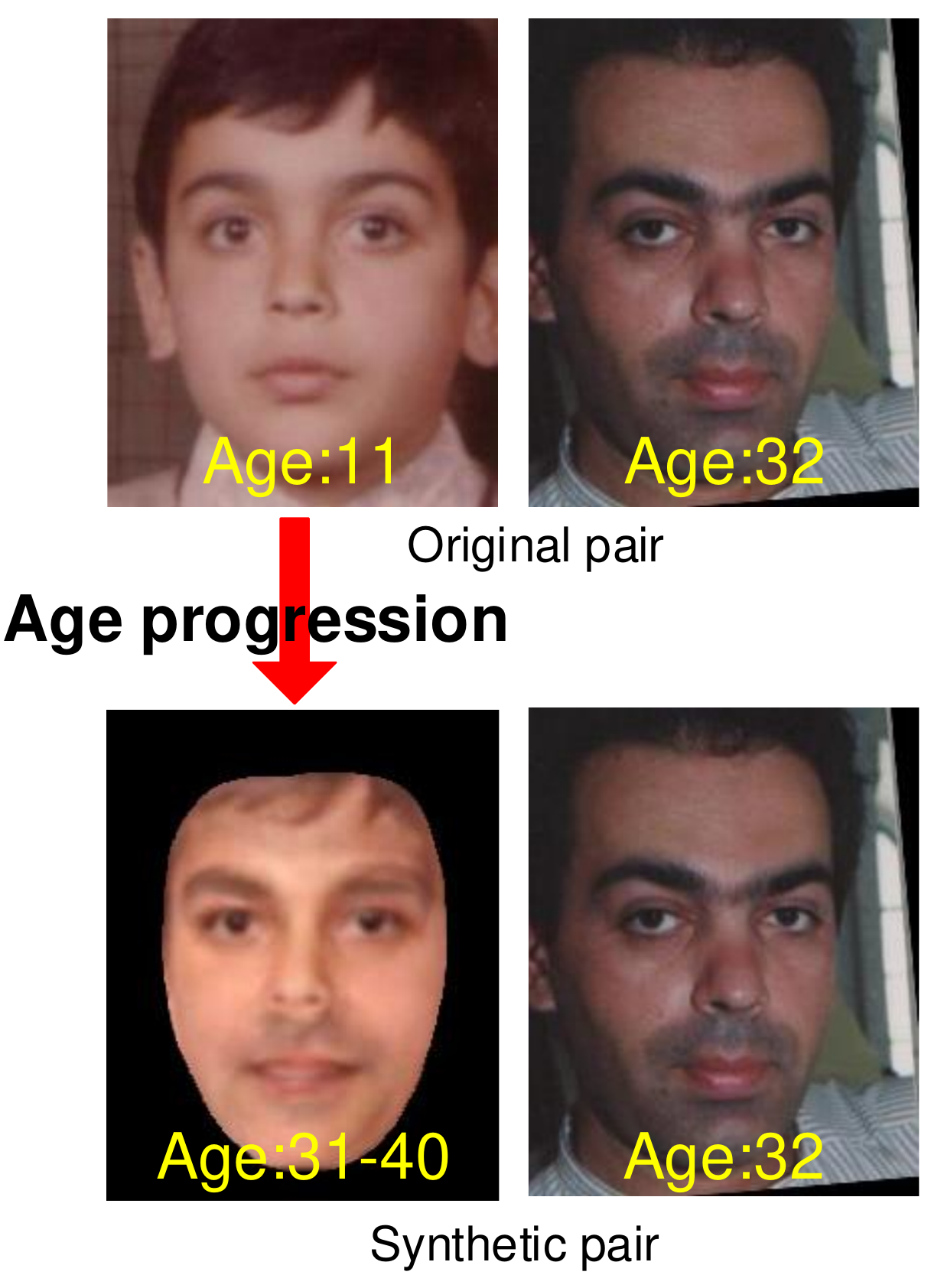}
		\label{fig6a}
	}
	\subfigure[FAR-FRR curve.]{
		\includegraphics[scale=0.32]{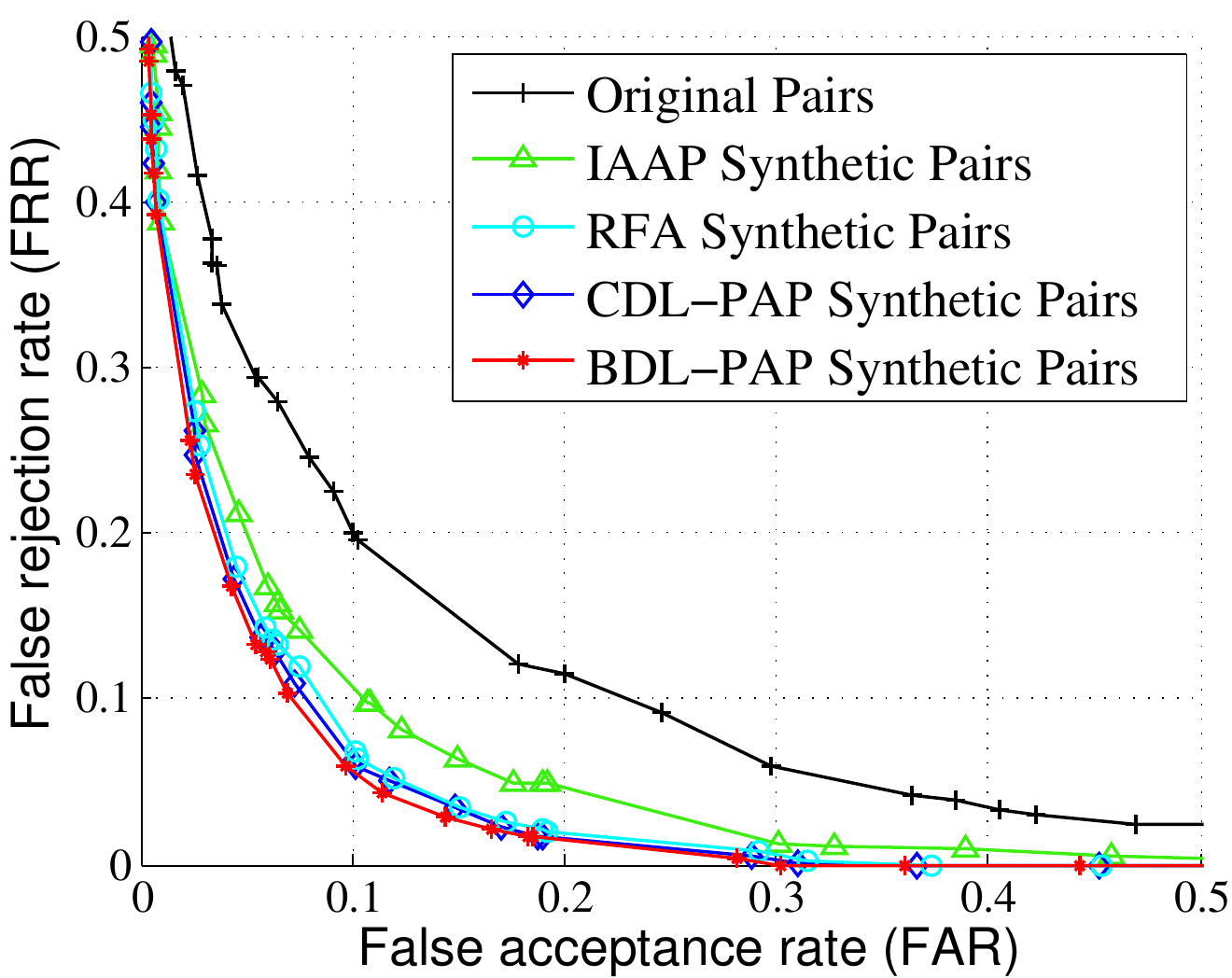}
		\label{fig6b}
	}
	\caption{Pair setting and comparative performance of face verification.}
	\label{fig6}
\end{figure}

To validate the improved performance of cross-age face verification with the help of the proposed BDL-PAP, we prepare for the intra-person pairs and inter-person pairs with cross ages on the FG-NET dataset. By removing undetected face photos and face pairs with age span no more than 20 years, we select 1,832 pairs (916 intra-person pairs and 916 inter-person pairs), called ``Original Pairs". Among the 1,832 pairs, we render the younger face in each pair to the aging face with the same age of the older face by the proposed BDL-PAP. Replacing each younger face with the corresponding aging face, we newly construct 1,832 pairs of the aging face and older face, called ``BDL-PAP Synthetic Pairs". Figure~\ref{fig_show_pair} shows the comparisons of and the original face pairs and the synthetic pairs by the proposed method, and Figure~\ref{fig6a} shows the pair setting. To evaluate performance of the proposed BDL-PAP, we also prepare the ``IAAP Synthetic Pairs",  ``RFA Synthetic Pairs", and ``CDL-PAP Synthetic Pairs" by IAAP ~\cite{kemelmacher2014illumination}, Recurrent Face Aging (RFA) ~\cite{wang2016recurrent}, and CDL-PAP~\cite{shu2015personalized}, respectively.

\begin{table*} [!th]\centering
	\renewcommand\arraystretch{1.5}
	\caption{Equal error rates (EER) (\%) of cross-age face verification.}
	\label{table_face_verification}
	\begin{tabular}{|c|c|c|c|c|c|}	
		\hline			
		\multirow{2}{*}{Pair settings} &\multirow{2}{*}{ Original Pairs}&{ IAAP }&{ RFA}&{ CDL-PAP }&{BDL-PAP } \\
		&
		{}&{Synthetic Pairs}&{Synthetic Pairs}&{Synthetic Pairs}&{Synthetic Pairs}\\
		\hline
		{EER (\%)} &{14.89}&{10.36}&{8.69}&{8.53}&{\bf 8.06}\\
		\hline 	
	\end{tabular}
\end{table*}

The detailed implementation of face verification is given as follows. First, we formulate a face verification model with deep Convolutional Neural Networks (deep ConvNets), which is based on DeepID2~\cite{sun2014deepID2}. Since we focus on the age progression in this work, please refer to~\cite{sun2014deepID2,taigman2014deepface} for more details of face verification with deep ConvNets. Second, we train this face verification model on the LFW dataset~\cite{LFWTech}, which is constructed for face verification. Third, we test the face verification on different face pairs.

The False Acceptance Rate - False Rejection Rate (FAR-FRR) curves and the Equal Error Rates (EER) on original pairs and synthetic pairs are shown in Figure~\ref{fig6}(b) and Table~\ref{table_face_verification}, respectively. We can see that the face verification on BDL Synthetic Pairs achieves the lower ERR than on Original Pairs, IAAP Synthetic Pairs, RFA Synthetic Pairs, and  CDL-PAP Synthetic Pairs. This illustrates that the aging faces by the proposed age progression method can effectively mitigate the effect of the age gap in cross-age face verification. The results also validate that, for a given input face, BDL-PAP can render a personalized and authentic aging face closer to the ground truth than the other compared methods.

\subsection{Effect of Personalized Layer}
To demonstrate the superiority of the personalization introduced by the proposed BDL-PAP, we conduct both qualitative and quantitative comparisons of BDL-PAP and its unpersonalized version BDL-AP on FG-NET. The proposed BDL-PAP can be degenerated into BDL-AP by the following two steps: 1) in the optimization procedure of Eq.~\eqref{eq5.1}, we set ${\bf P}^g$ to be a full-zero matrix when initializing and updating it in each iteration; 2) in the synthesis of age progression, we directly set ${\bf p}^g$ to be a full-zero vector in Eq.~\eqref{eq61}.

\begin{figure}[t]
	\centering
	\includegraphics[scale=0.26]{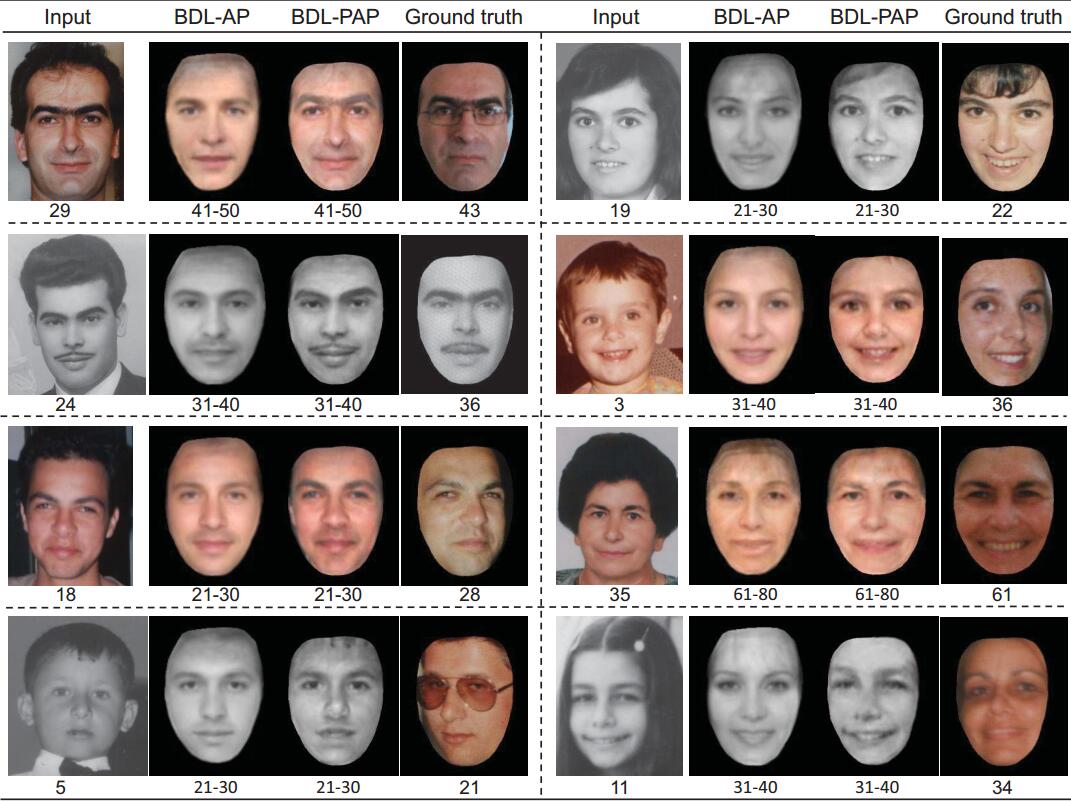}
	\caption{The comparisons of BDL-PAP and BDL-AP on FG-NET.}
	\label{fig_without_p}
\end{figure}

We first qualitatively compare the experimental results of BDL-PAP and BDL-AP, which are illustrated in  Figure~\ref{fig_without_p}. We can see that the aging results generated by BDL-PAP are more similar to the ground truth than the counterparts of BDL-AP. In particular, compared with BDL-AP, BDL-PAP can preserve much more personalized facial characteristics, such as the special eyebrow shape in the top-left face and the mole in the fifth face of the first row, which are preserved by BDL-PAP but discarded by BDL-AP.

We also use the two-way ANalysis Of VAriance (ANOVA)~\cite{Minium2003Statistical} test to quantitatively compare BDL-PAP and BDL-AP. To avoid bias as much as possible, we invite 30 users covering a wide range of age and from all walks of life to browse 50 comparison groups, each of which includes the aging results of BDL-PAP and BDL-AP, the input face and the ground truth. Two aging faces in each comparison group are shown to users with a random order. For each comparison, if one user thinks the two results are comparable with each other, BDL-PAP and BDL-AP are assigned with score 1 respectively; if the user thinks one aging result is better than the other one, the method corresponding to the better aging result is assigned with score 2, and the other method is assigned with score 0. The comparison results are given in Table~\ref{table_ANOVA}. We can see that BDL-PAP gets a score $1.5779$, which is much higher than the score $0.4221$ obtained by BDL-AP. This result validates the effectiveness of bringing in the  personalization. Moreover, the $p$-values of ANOVA test show that this superiority is statistically significant and the difference of the users is insignificant.

\begin{table*} [!th]\centering
	{ \caption{The left part illustrates the average rating scores and standard deviation values from the user study on the comparisons of BDL-PAP and BDL-AP . The right part shows the ANOVA test results.
		}
		\label{table_ANOVA}
		{
			\begin{tabular}{|c|c||c|c|c|c|}
				\hline
				\multicolumn{2}{|c||}{BDL-PAP {vs.} BDL-AP } &\multicolumn{2}{c|}{Factor of approaches}&\multicolumn{2}{c|}{Factor of users} \\  				
				\cline{1-6}
				{     BDL-PAP       }&{BDL-AP    } &{$F$-statistic} &{$p$-value}&{$F$-statistic} &{$p$-value}\\
				\hline{${\bf 1.5779 \pm 0.0363
						}$} &{$0.4221 \pm 0.0363$} &{$128.6381$} &{$4.10 \times 10^{-8}$}&{$1.88\times 10^{-15}$} &{$1.0000$}\\
				\hline
			\end{tabular}
		}}
	\end{table*}
	
\begin{table*}[!t]
	\centering
	\caption{Comparison of running time for age progression synthesis.}
	\label{tab_run}
	\begin{tabular}{c}
		\centering
		\includegraphics[scale=0.84]{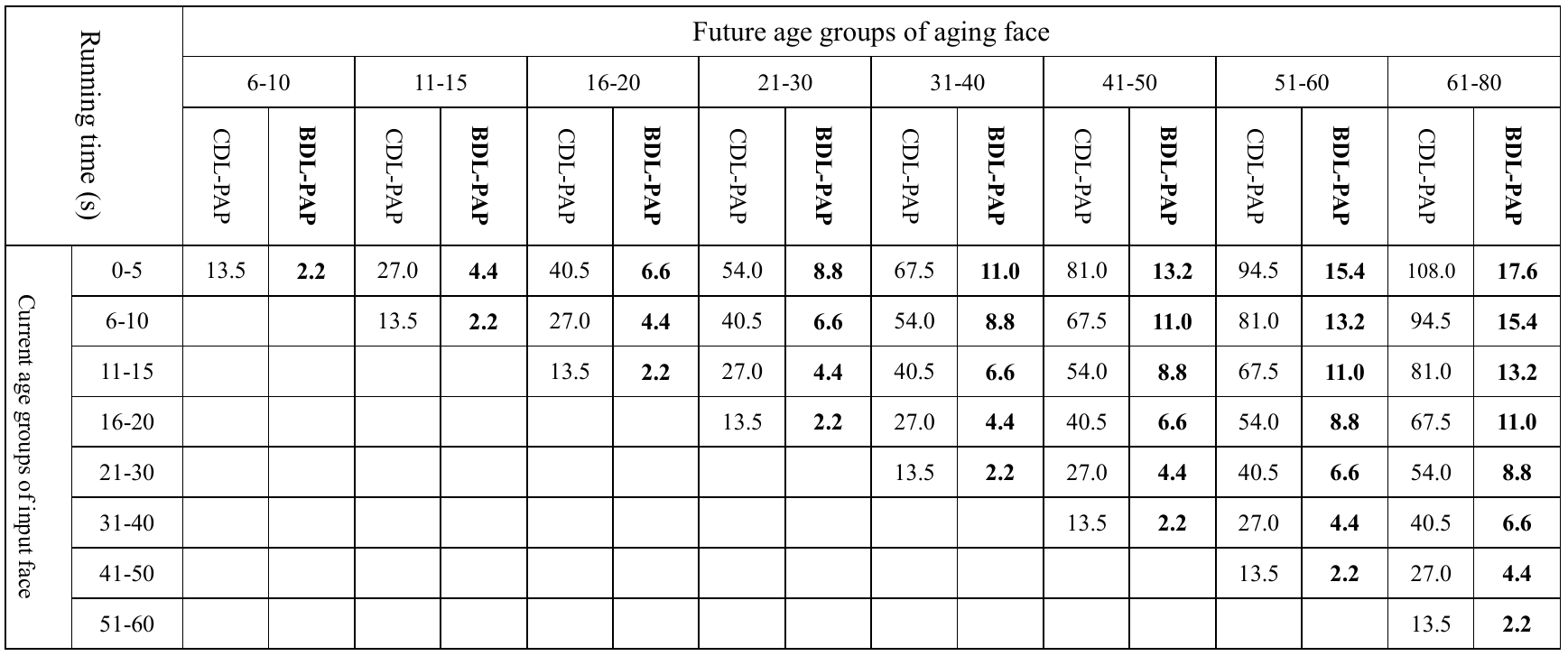}
	\end{tabular}
\end{table*}

\subsection{Comparison of running time}
Another advantage of BDL-PAP lies in the significant reduction of the running time of age progression synthesis compared with CDL-PAP~\cite{shu2015personalized}. To illustrate this advantage of BDL-PAP, we test the running time of BDL-PAP and CDL-PAP on a PC with CPU Intel Core i7 3.6 GHz and 6 GB memory. Table~\ref{tab_run} shows the comparison of the running time. When we generate an aging face in the ``6-10" age group from an input face in the ``0-5" age group, BDL-PAP and CDL-PAP take 2.2s and 13.5s, respectively. And when we generate an aging face in the ``61-80" age group from the input face in the ``0-5" age group, BDL-PAP and CDL-PAP take 17.6s and 108.0s, respectively. We can see that BDL-PAP is much more efficient than CDL-PAP, and the running time of BDL-PAP and CDL-PAP increases accompanied with the increment of the age  difference between the input face and the aging face. Moreover, we compare the running time of IAAP~\cite{kemelmacher2014illumination} and BDL-PAP. Since FT Demo requires a series of manual operations, we do not compare its running time. Given an input face in the ``0-5" age group, generating its aging face sequence covering all the future age groups needs about 72.0s and 17.6s by IAAP and BDL-PAP, respectively. Thus, BDL-PAP is also much faster than IAAP.

\section{Conclusions and Future Work}
\label{C}
In this work, we proposed a personalized age progression method. Basically, we design multiple aging dictionaries for different age groups, in which the dictionary bases from two neighboring dictionaries respectively form a particular aging process pattern across different age groups, and a linear combination of these patterns expresses a particular aging process. Moreover, we define the aging layer and the personalized layer for an individual to capture the aging characteristics and the personalized characteristics, respectively. We train all aging dictionaries on the collected short-term aging database. Specifically, the younger- and older- age face pairs of the same persons are used to train two aging dictionaries in the neighboring age groups with the common sparse representation, excluding the specific personalized layer. Given a face, we render its aging face sequence from the current age to the future age step by step on the learned aging dictionaries.
In future, we consider the face anti-aging synthesis, namely restoring the younger face for a given older face.


		\section{Acknowledgments}
This work was partially supported by the 973 Program (Project No. 2014CB347600), the National Natural Science Foundation of China (Grant No. 61522203, 61572252 and 61672285), the Natural Science Foundation of Jiangsu Province (Grant No. BK20140058 and BK20150755), and the National Ten Thousand Talent Program of China (Young Top-Notch Talent). Jinhui Tang is the corresponding author.


\small{
	\bibliographystyle{IEEEtran}
	\bibliography{egbib}
}

\begin{IEEEbiography}[{\includegraphics[width=1in,height=1.25in,clip,keepaspectratio]{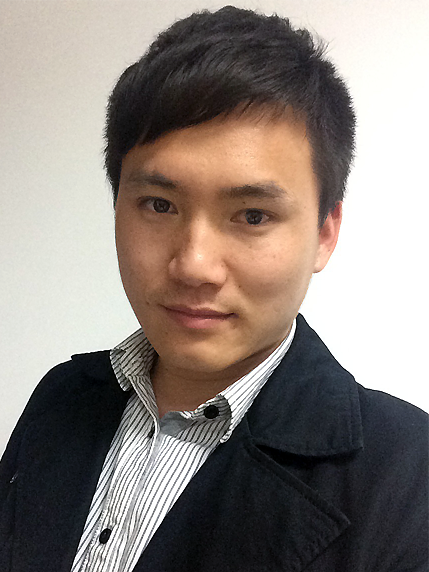}}]{Xiangbo Shu}
	is an Assistant Professor in School of Computer Science and Engineering, Nanjing University of Science and Technology, China. He received his Ph.D. degree in July 2016 from Nanjing University of Science and Technology. From 2014 to 2015, he worked as a visiting scholar in the Department of Electrical and Computer Engineering at National University of Singapore. His research interests include computer vision, and machine learning. He has received the Best Student Paper Award in MMM 2016 and the Best Paper Runner-up in ACM MM 2015.
\end{IEEEbiography}

\begin{IEEEbiography}[{\includegraphics[width=1in,height=1.25in,clip,keepaspectratio]{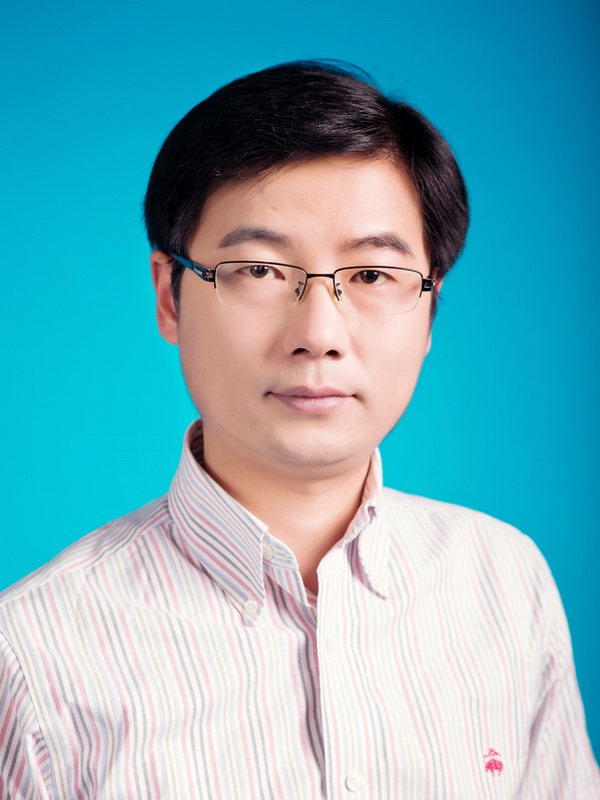}}]{Jinhui Tang}
is a Professor in School of Computer Science and Engineering, Nanjing University of Science and Technology, China. He received his B.E. and Ph.D. degrees in July 2003 and July 2008 respectively, both from the University of Science and Technology of China. From 2008 to 2010, he worked as a research fellow in School of Computing, National University of Singapore. His current research interests include large-scale multimedia search. He has authored over 100 journal and conference papers in these areas. Prof. Tang is a recipient of ACM China Rising Star Award and a co-recipient of the Best Student Paper Award in MMM 2016, and Best Paper Award in ACM MM 2007, PCM 2011 and ICIMCS 2011.
\end{IEEEbiography}

\begin{IEEEbiography}[{\includegraphics[width=1in,height=1.25in,clip,keepaspectratio]{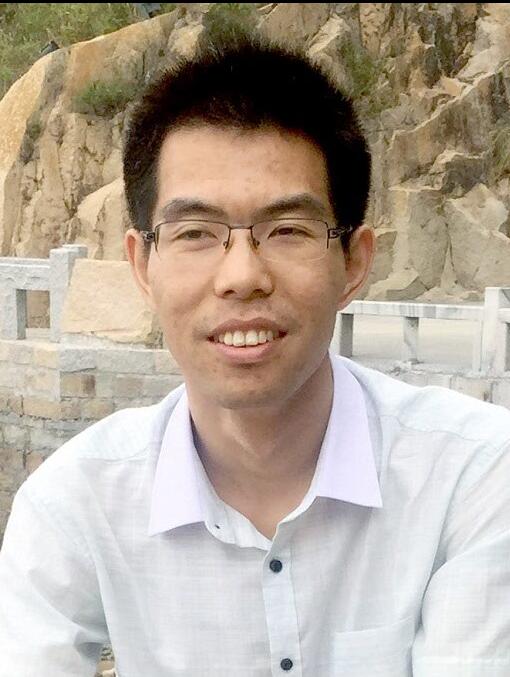}}]{Zechao Li}
	is an Associate Professor in School of Computer Science and Engineering, Nanjing University of Science and Technology. He received the Ph.D degree from Institute of Automation, Chinese Academy of Sciences, China, in 2013, and the B.E. degree from University of Science and Technology of China (USTC), China, in 2008. His research interests include large-scale multimedia understanding, social media mining, subspace learning, etc. He received the 2015 Excellent Doctoral Dissertation of Chinese Academy of Sciences, the 2015 Excellent Doctoral Dissertation of China Computer Federation, the Top 10\% Paper Award of IEEE MMSP 2015 and the 2013 President Scholarship of Chinese Academy of Science.
\end{IEEEbiography}

\begin{IEEEbiography}[{\includegraphics[width=1in,height=1.25in,clip,keepaspectratio]{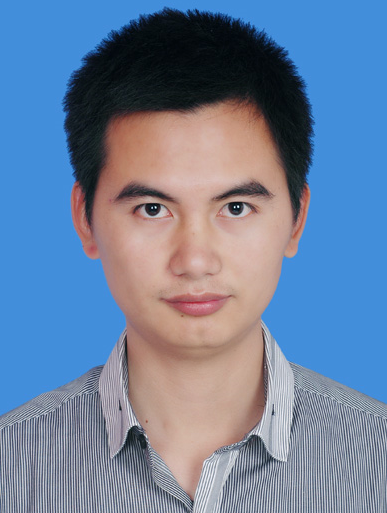}}]{Hanjiang Lai}
	received his B.S. and Ph.D. degrees
	from Sun Yat-sen University in 2009 and 2014,
	respectively. He was working as a research fellow
	at National University of Singapore during 2014-2015. He is now working at Sun Yat-sen university.
	His research interests includes machine learning
	algorithms, deep learning, and computer vision.
\end{IEEEbiography}

\begin{IEEEbiography}[{\includegraphics[width=1in,height=1.25in,clip,keepaspectratio]{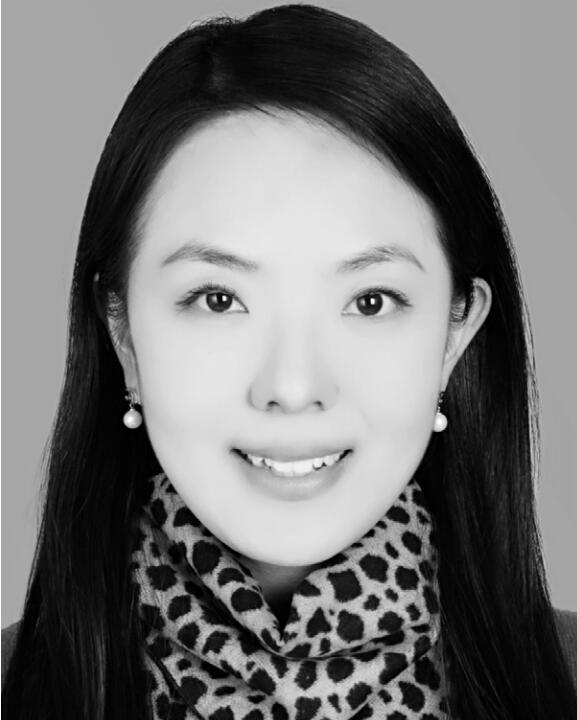}}]{Liyan Zhang}
	received the Ph.D. degree in computer
	science  from  the  University  of  California,  Irvine,
	in 2014. She is currently an Associate Professor with
	the  School  of  Computer  Science  and  Technology,
	Nanjing University of Aeronautics and Astronautics.
	Her research  interests  include multimedia analysis, and
	computer vision. She has
	received the Best Paper Award in ICMR 2013 and the Best Student Paper Award in MMM 2016.
\end{IEEEbiography}

\begin{IEEEbiography}[{\includegraphics[width=1in,height=1.25in,clip,keepaspectratio]{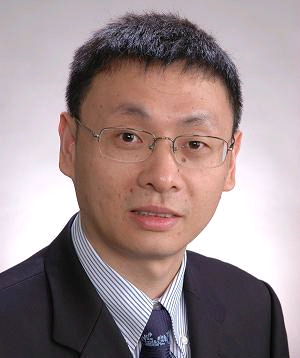}}]{Shuicheng Yan}
 is the Dean's Chair Associate Professor at National University of Singapore, and also the chief scientist of Qihoo/360 company.  Dr. Yan's research areas include machine learning, computer vision and multimedia, and he has authored/co-authored hundreds of technical paper, with Google Scholar citation $>$15,000 times and H-index 52. He has been serving as an associate editor of IEEE TCSVT and ACM TIST. He received the Best Paper Awards from ACM MM'13 (Best Paper and Best Student Paper), MMM'16 (Best Student Paper), ACM MM'12 (Best Demo), PCM'11, ACM MM'10, ICME'10 and ICIMCS'09, the winner prizes of the classification task in PASCAL VOC 2010-2012, 2011 Singapore Young Scientist Award, and 2012 NUS Young Researcher Award. He is also IEEE and IAPR Fellow.
\end{IEEEbiography}
\end{document}